\pdfoutput=1

\documentclass[11pt]{article}

\usepackage[final]{acl}

\usepackage{times}
\usepackage{latexsym}

\usepackage[T1]{fontenc}

\usepackage[utf8]{inputenc}

\usepackage{microtype}

\usepackage{inconsolata}

\usepackage{tabularx}

\usepackage{url}
\usepackage{graphicx}
\usepackage{amsmath}
\usepackage{multirow}
\usepackage{booktabs}
\usepackage{listings}
\usepackage{wasysym}
\usepackage{amssymb}
\usepackage{mdframed}
\usepackage{authblk}
\usepackage{xcolor}
\usepackage{tcolorbox}
\tcbuselibrary{listings,breakable}
\usepackage{makecell}
\usepackage{pifont}

\usepackage{enumitem}
\usepackage{graphicx}

\newcommand{\cofirst}{$^\ast$}
\newcommand{\corresponding}{\textsuperscript{$\dagger$}}

\author[1\cofirst]{\bf Qinzhuo Wu}
\author[12\cofirst]{\bf Zhizhuo Yang}
\author[13]{\bf Hanhao Li}
\author[1\corresponding]{\bf Pengzhi Gao}
\author[1]{\bf Wei Liu}
\author[1]{\bf Jian Luan}
\affil[ ]{\textsuperscript{1}MiLM Plus, Xiaomi Inc \qquad 
          \textsuperscript{2}Peking University \qquad
          \textsuperscript{3}Chinese University of Hong Kong}
\affil[ ]{\footnotesize \texttt{\{wuqinzhuo, gaopengzhi, liuwei40, luanjian\}@xiaomi.com},}
\affil[ ]{\footnotesize \texttt{\{yzz2022\}@stu.pku.edu.cn},\quad\texttt{\{lihanhao\}@link.cuhk.edu.hk}}
\title{MobileBench-OL: A Comprehensive Chinese Benchmark for Evaluating Mobile GUI Agents in Real-World Environment}

\begin{document}
\maketitle
\let\thefootnote\relax\footnotetext{\cofirst\ Equal contribution.}
\let\thefootnote\relax\footnotetext{\corresponding\ Corresponding authors.}
\begin{abstract}
Recent advances in mobile Graphical User Interface (GUI) agents highlight the growing need for comprehensive evaluation benchmarks. While new online benchmarks offer more realistic testing than offline ones, they  tend to focus on the agents' task instruction-following ability while neglecting their reasoning and exploration ability. Moreover, these benchmarks do not consider the random noise in real-world mobile environments. This leads to a gap between benchmarks and real-world environments.
To addressing these limitations, we propose MobileBench-OL, an online benchmark with 1080 tasks from 80 Chinese apps. It measures task execution, complex reasoning, and noise robustness of agents by including 5 subsets, which set multiple evaluation dimensions.
We also provide an auto-eval framework with a reset mechanism, enabling stable and repeatable real-world benchmarking.
Evaluating 12 leading GUI agents on MobileBench-OL shows significant room for improvement to meet real-world requirements. Human evaluation further confirms that MobileBench-OL can reliably measure the performance of leading GUI agents in real environments.
Our data and code will be released upon acceptance.

\end{abstract}

\begin{figure}[!t]
\centering
\includegraphics[width=0.99\columnwidth]{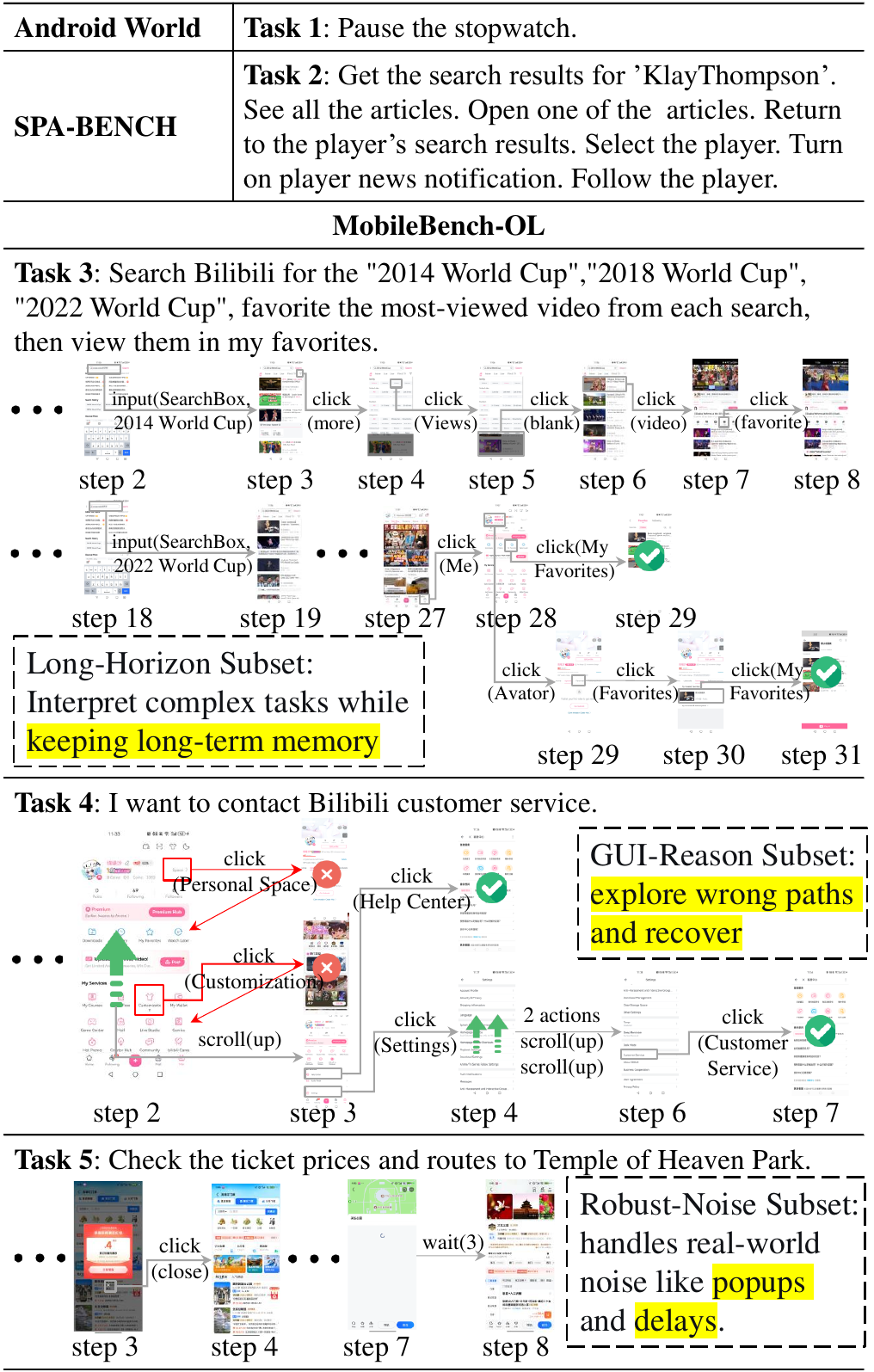} 
\caption{In addition to instruction-following ability, MobileBench-OL also measures long-horizon reasoning, exploration, and real-world noise handling. }
\label{figure1_1}
\end{figure}

\begin{table*}[!t]
    \centering
\resizebox{1\textwidth}{!}
{
\begin{tabular}{l|cccccccccc}
\hline
\multirow{2}{*}{Benchmark}        & \multirow{2}{*}{Tasks}  &  \multirow{2}{*}{Apps}  &   \!\!\multirow{2}{*}{Dynamic}\!\!  &   \!\!Third-party\!\!  &  \!\! Chinese \!\!  & Difficulty & Long- & GUI-  & Noise- &   \!\!\multirow{2}{*}{Reset}\!\! \\
& & & &  Apps & Apps & Levels &  Horizon  &   \!\!Reasoning \!\!  & Robust & \\ \hline
AITW  \cite{rawles2023androidwildlargescaledataset}           & 30378 & 357  & \color{red}{\ding{55}}      & \color{green}{\ding{51}}           & \color{red}{\ding{55}}   & \color{red}{\ding{55}}         & \color{red}{\ding{55}}                 & \color{red}{\ding{55}}               & \color{red}{\ding{55}}     & \color{red}{\ding{55}}  \\
AITZ    \cite{zhang-etal-2024-android}         & 2504  & 70   & \color{red}{\ding{55}}       & \color{green}{\ding{51}}           & \color{red}{\ding{55}}    & \color{red}{\ding{55}}        & \color{red}{\ding{55}}                 & \color{red}{\ding{55}}               & \color{red}{\ding{55}}    & \color{red}{\ding{55}}   \\
AndroidControl \cite{li2024effects}   & 15283 & 833  & \color{red}{\ding{55}}       & \color{green}{\ding{51}}       & \color{red}{\ding{55}}      & \color{red}{\ding{55}}          & \color{red}{\ding{55}}                 & \color{red}{\ding{55}}               & \color{red}{\ding{55}}    & \color{red}{\ding{55}}   \\
MobileAgentBench \cite{wang2024mobileagentbench} \!\! & 100   & 10   & \color{red}{\ding{55}}       & \color{red}{\ding{55}}   & \color{red}{\ding{55}}      & \color{green}{\ding{51}}      & \color{red}{\ding{55}}          & \color{red}{\ding{55}}                 & \color{red}{\ding{55}}               & \color{red}{\ding{55}}     \\
AppAgent    \cite{zhang2023appagentmultimodalagentssmartphone}     & 50    & 10   & \color{red}{\ding{55}}       & \color{green}{\ding{51}}           & \color{red}{\ding{55}}          & \color{red}{\ding{55}}                 & \color{red}{\ding{55}}     & \color{red}{\ding{55}}            & \color{red}{\ding{55}}   & \color{red}{\ding{55}}    \\
LearnGUI-Offline \cite{liu2025learnact} & 2253  & 73   & \color{red}{\ding{55}}       & \color{green}{\ding{51}}           & \color{red}{\ding{55}}          & \color{red}{\ding{55}}                 & \color{red}{\ding{55}}       & \color{red}{\ding{55}}          & \color{red}{\ding{55}}    & \color{red}{\ding{55}}   \\ 
GUI-Robust \cite{yang2025gui}  & 5318   &  392   &  \color{red}{\ding{55}}      & \color{green}{\ding{51}}           & \color{green}{\ding{51}}          & \color{red}{\ding{55}}                 & \color{red}{\ding{55}}     & \color{red}{\ding{55}}            & \color{green}{\ding{51}} & \color{red}{\ding{55}}     \\\hline
AndroidArena   \cite{xing2024understanding}  & 221   & 4    & \color{green}{\ding{51}}       & \color{red}{\ding{55}}           & \color{red}{\ding{55}}          & \color{red}{\ding{55}}                 & \color{red}{\ding{55}}       & \color{red}{\ding{55}}          & \color{red}{\ding{55}}   & \color{red}{\ding{55}}    \\
LLamaTouch    \cite{zhang2024llamatouch}   & 496   & 57   & \color{green}{\ding{51}}       & \color{green}{\ding{51}}           & \color{red}{\ding{55}}          & \color{red}{\ding{55}}                 & \color{red}{\ding{55}}       & \color{red}{\ding{55}}          & \color{red}{\ding{55}}    & \color{red}{\ding{55}}   \\
AndroidWorld \cite{rawles2024androidworld}    & 116   & 20   & \color{green}{\ding{51}}       & \color{green}{\ding{51}}      & \color{red}{\ding{55}}     & \color{green}{\ding{51}}    & \color{red}{\ding{55}}          & \color{red}{\ding{55}}                 & \color{red}{\ding{55}}               & \color{red}{\ding{55}}     \\
AndroidLab   \cite{xu2024androidlab}    & 138   & 9    & \color{green}{\ding{51}}       & \color{green}{\ding{51}}           & \color{red}{\ding{55}}          & \color{red}{\ding{55}}                 & \color{red}{\ding{55}}        & \color{red}{\ding{55}}         & \color{red}{\ding{55}}    & \color{red}{\ding{55}}   \\
A3      \cite{chai2025a3}         & 201   & 20   & \color{green}{\ding{51}}       & \color{green}{\ding{51}}       & \color{red}{\ding{55}}    & \color{green}{\ding{51}}    & \color{red}{\ding{55}}          & \color{red}{\ding{55}}                 & \color{red}{\ding{55}}               & \color{red}{\ding{55}}     \\
B-Moca     \cite{lee2024benchmarking}      & 131   & 15   & \color{green}{\ding{51}}       & \color{red}{\ding{55}}           & \color{red}{\ding{55}}          & \color{red}{\ding{55}}                 & \color{red}{\ding{55}}      & \color{red}{\ding{55}}           & \color{red}{\ding{55}}    & \color{red}{\ding{55}}   \\
LearnGUI-Online \cite{liu2025learnact}  & 101   & 23   & \color{green}{\ding{51}}       & \color{red}{\ding{55}}           & \color{red}{\ding{55}}          & \color{red}{\ding{55}}                 & \color{red}{\ding{55}}     & \color{red}{\ding{55}}            & \color{red}{\ding{55}}  & \color{red}{\ding{55}}     \\
D-GARA \cite{chen2025d}  & 152   &  7   & \color{green}{\ding{51}}       & \color{green}{\ding{51}}           & \color{green}{\ding{51}}          & \color{red}{\ding{55}}                 & \color{red}{\ding{55}}     & \color{red}{\ding{55}}            & \color{green}{\ding{51}} & \color{red}{\ding{55}}     \\
SPA-Bench  \cite{chen2024spa}      & 340   & 66   & \color{green}{\ding{51}}       & \color{green}{\ding{51}}   & \color{green}{\ding{51}}      & \color{green}{\ding{51}}    & \color{green}{\ding{51}}          & \color{red}{\ding{55}}                 & \color{red}{\ding{55}}               & \color{red}{\ding{55}}     \\ \hline
MobileBench-OL   & 1080   & 80   & \color{green}{\ding{51}}       & \color{green}{\ding{51}}           & \color{green}{\ding{51}}          & \color{green}{\ding{51}}       & \color{green}{\ding{51}}          & \color{green}{\ding{51}}      & \color{green}{\ding{51}}           & \color{green}{\ding{51}}    \\\hline 
\end{tabular}}
    \caption{Comparison of different datasets and environments for benchmarking Mobile GUI agents.}
    \label{table_1}
\end{table*}

\section{Introduction}

In recent years, the research community has shown growing interest in developing mobile GUI agents \cite{hong2023cogagent,wu2025reachagent}, which have demonstrated remarkable performance in controlling mobile devices.
As GUI agents continue to emerge, the fair evaluation of their performance becomes essential, leading to the growing need for benchmarks. 
Previous offline benchmarks \cite{chen2024gui,rawles2023androidwildlargescaledataset,li2024effects} assess agents in static GUI environments. Actions are generated from static screenshots and then compared against predefined standard answers.
These methods ignore the impact of unexpected factors in real-world environments, such as pop-ups, and cannot handle scenarios where a single task may have multiple valid GUI trajectories. 
Therefore, some studies \cite{xie2024osworld,lee2024benchmarking,zhang2023mobile} have introduced online evaluation benchmarks that deploy apps in device emulators to simulate real-world environments. They use device state \cite{rawles2024androidworld} or LLM scores \cite{sun2025autoeval} to evaluate the generated GUI trajectories. 
Unlike offline methods, these online benchmarks assess agent performance based on task completion rather than a predefined standard trajectory, offering a more accurate measure of performance in dynamic environments.

Despite significant progress, previous online benchmarks still have several limitations:
First, most benchmarks focus on simple, atomic tasks or rigid, step-by-step instructions. They do not assess an agent’s ability to reason and explore autonomously in dynamic environments.
In Figure \ref{figure1_1}, simple tasks like "Pause the stopwatch" test only core app functions, while rigid tasks like Task 2 provide detailed steps that evaluate instruction-following but lead to fixed solutions. 
In contrast, a comprehensive benchmark should include complex, open-ended tasks that require long-term planning and GUI reasoning. Long-horizon tasks like "sequentially search for and favorite popular videos related to three World Cups" measure the ability to maintain focus on a high-level goal while executing multiple subtasks, allowing for diverse solution paths. 
Meanwhile, tasks like "contact customer service" examine GUI reasoning abilities. Since customer service functionality is hidden deep within the interface, the agent needs to first explore semantically related areas (e.g., Personal Space and Customization), then recover from incorrect paths. 
Beyond simply following instructions, such a benchmark also tests an agent’s ability to interpret tasks and autonomously explore interfaces to complete them.
Furthermore, existing benchmarks overlook the unpredictability of real-world mobile environments, such as unexpected pop-ups, interface lag, or operation errors. In Figure \ref{figure1_1}, an ad pop-up requires clicking accurately on the close icon to proceed, and a network delay page requires agent to wait a few seconds. Current benchmarks fail to measure how well agents handle such real-world variability, creating a gap between academic evaluation and real-world deployment.


\begin{table*}[!t]
    \centering
\resizebox{1\textwidth}{!}{
    \begin{tabular}{c|l|c|c|c|c|l}
    \hline\hline
    \multirow{2}{*}{Subset}  & \makecell[c]{\multirow{2}{*}{Task Examples}} & \!\!\!\!Golden\!\!\!\! & \!\!\!\!Functional\!\!\!\! & \!\!\multirow{2}{*}{Weight}\!\! & \multirow{2}{*}{Difficulty} & \makecell[c]{\multirow{2}{*}{Challenges}} \\  
    & & Steps & Point & & & \\ \hline
    Base     &  Search for Pokemon on Bilibili. & 4 & Search & - & Easy (step:4)  & Covering vairous functional points\!\!\!\! \\\hline
    \multirow{2}{*}{\!\!Long-Tail\!\!} &  Set my main team to Premier League‘s  & \multirow{2}{*}{8} & Personal & \multirow{2}{*}{-} & \!\!\!\!\multirow{2}{*}{Medium (step:8)}\!\!\!\!   & Basic tasks with novel UIs
in  \\
 & Arsenal in Dongqiudi. & & Center & & & long-tail apps.\\ \hline 
  & Open Tomato Novel, search for "War and   & \multirow{4}{*}{31} & \multirow{4}{*}{Search} & \multirow{4}{*}{-} & \multirow{4}{*}{Hard (step:31)}   & \multirow{4}{*}{\makecell[l]{Hard tasks require more than\\20 steps to complete.}} \\
Long- & Peace","The Lady of the Camellias","The   &  & & & &   \\  
Horizon& Ordinary World","Tallow Ball" in order,  &  & & & \\  
& follow the authors of these four books. &  & & & \\ \hline 
 \multirow{3}{*}{\makecell[c]{GUI-\\\!\!Reasoning\!\!}} & Navigate to Toutiao’s scan function.  & 6 & Channel  & 0.5 & \!\!Easy (weight:0.5)\!\!   & Tasks require exploration and \\ \cline{2-6}
  & I want to contact Bilibili customer service.  & 4 & Settings  & 3 & Hard (weight:3)   & reasoning, e.g. Icon
Understanding,\!\!\!\! \\ \cline{2-6}
  & Set a 15-minute turn-off timer in Bilibili.   & 11 & Settings  & 4 & Hard (weight:4)   & Hidden
Function Discovery. \\ \hline
 Noise- & Go to DingTalk and send a picture from  & \multirow{2}{*}{9} & \multirow{2}{*}{Message} & \multirow{2}{*}{-} & \!\!\!\!\multirow{2}{*}{Medium (step:9)}\!\!\!\!    & Randomly introduce noise page  \\ 
 Robust &   the album to the DingTalk assistant.  & &   &   &    & during inference, e.g.
pop-ups. \\\hline\hline
    \end{tabular}}
    \caption{Example tasks for the five subsets. Note that Base, Long-Tail, Long-Horizon, and Noise-Robust subsets use the number of golden steps to classify difficulty levels, while the GUI-Reasoning subset uses exploration weight.}
    \label{table_5}
\end{table*}

To address these limitations, we propose MobileBench-OL, an online benchmark with 1080 real-world GUI tasks from 80 Chinese apps. It evaluates agents across three core dimensions via five subsets: (1) Base Capabilities, the Base and Long-Tail subsets; (2) Complex Reasoning, Long-Horizon ($\geq$20 steps) and GUI-Reasoning (requires exploration) subsets; (3) Robustness, Noise-Robust subset (4 artificial noises).
We also introduce an automated evaluation (Auto-Eval) framework that assesses trajectories without manual intervention or system data. A fine-grained, instruction-based Reset Mechanism returns devices to their initial state, ensuring stability and repeatability.


Our contributions can be summarized as follows: 
\begin{itemize}[itemsep=2pt,topsep=0pt,parsep=0pt]
\item  We propose MobileBench-OL, a comprehensive benchmark of 1080 real-world tasks across 80 apps. It includes five subsets and evaluates multiple dimensions, such as basic capabilities, complex reasoning, and robustness, to thoroughly assess agent performance in real-world environments.
\item  We propose an Auto-Eval framework with a Reset Mechanism that ensures stable and reproducible task generation and execution.
\item  Evaluations of 12 leading GUI Agents reveal significant room for improvement in real-world performance. Human evaluation further confirms the benchmark provides stable and reliable measurements.
\end{itemize}

\section{Related Work}
Table \ref{table_1} shows the comparison of MobileBench-OL with other benchmarks.
With the rapid advancement of GUI Agents, traditional static benchmarks \cite{wu2024mobilevlm, zhan2023you,deng2023mind2web} have become inadequate.
Recently proposed dynamic environments such as AndroidWorld \cite{rawles2024androidworld}, LlamaTouch \cite{zhang2024llamatouch}, and Mobile-Env \cite{zhang2023mobile} evaluate task completion via device state or LLM judgments rather than verifying trajectory–goal consistency.
However, many benchmarks rely on offline, static apps that differ from mainstream designs and therefore do not represent real-world usage \cite{liu2025llm}. While SPA-Bench \cite{chen2024spa} incorporates popular open-source apps, it cannot support apps and tasks that require account login or have high memory usage. Tasks in existing benchmarks are either too simple or overly rigid, neglecting the reasoning and exploration abilities  in dynamic environments.
Regarding real-world anomalies, GUI-Robust \cite{yang2025gui} proposes a static benchmark with seven anomaly scenarios. D-GARA \cite{chen2025d} introduces a dynamic benchmark in which anomaly interruptions are manually configured per specific task. We categorize four common noise types and randomly introduce them during inference, thereby evaluating the agent’s robustness against unpredictable environmental noise.
See Appendix \ref{appendix_related} for more related works.

\section{MobileBench-OL}
\subsection{Apps Selection}
To ensure authenticity and comprehensiveness, we divided apps into 12 categories based on the App Store's classification.
For the four subsets excluding the Long-Tail, we selected the most popular app from each category.
For the Long-Tail subset, we selected 5–6 apps per category with diverse core functions, totaling 68 apps.
All categories and icons are listed in Figure \ref{figure_appendix_1} and Table \ref{appendix_apps}.
Unlike earlier studies, we did not exclude apps that require account login or high memory usage, as our benchmarks were run on real devices.

\begin{figure*}[!tb]
\setlength{\abovecaptionskip}{0.2cm}
\setlength{\belowcaptionskip}{-0.1cm}
\centering
\includegraphics[width=1\textwidth]{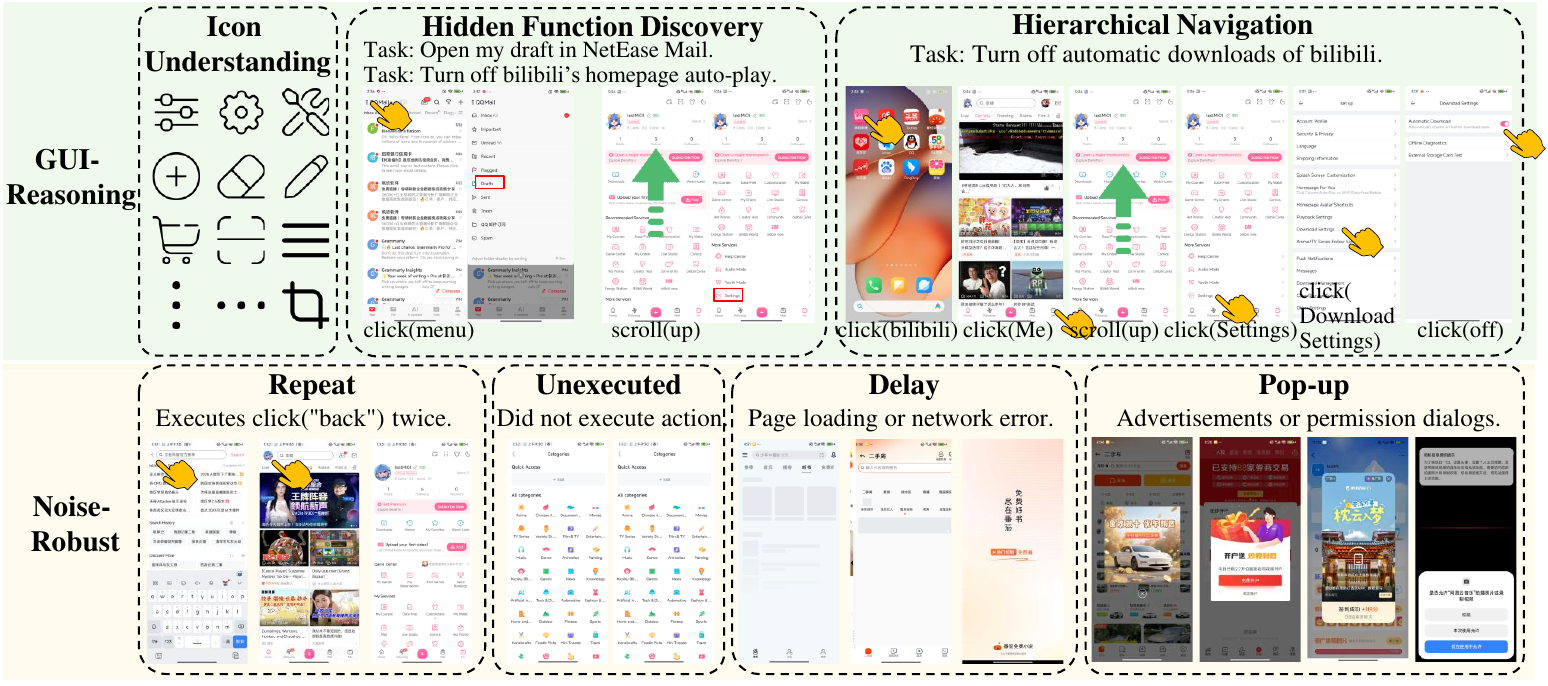} 
\caption{ Examples of three exploration abilities (above)  and four noise types (bottom). }
\label{figure3_2}
\end{figure*}

\subsection{Benchmark Subsets}
\subsubsection{Base Subset and Long-Tail Subset}
The Base and Long-Tail Subsets assess the agent’s basic abilities.
The Base Subset covers varied difficulty and function points from popular apps, while the Long-Tail Subset focuses on 68 less common apps.
Figures \ref{figure7_3} and \ref{figure7_4} in the Appendix show two example tasks. They both involve finding a product in a shopping app and navigating to reviews or the store homepage.
Though the high-level steps are similar, their UI designs and app-specific elements differ significantly.
Thus, the Long-Tail Subset evaluates whether the agent can apply general skills to novel interfaces, which tests its true robustness and adaptability.

\textbf{Difficulty Levels Based on Golden Steps:}
The difficulty of four subsets (excluding GUI-Reasoning) is defined by golden steps, as in Table \ref{table_5}: Easy (< 8 steps), Medium (8–19 steps), Hard ($\geq$ 20 steps). Golden steps reflect the minimal GUI pages a human needs to complete the task. Base, Long-Tail, and Noise-Robust subsets contain Easy or Medium tasks; Long-Horizon tasks are all Hard.

\textbf{Functional Point Coverage:}
In Base subset, the tasks cover all of the app's key functions. 
In Figure \ref{figure3_2}, while Bilibili's primary function is keyword-based video "search," secondary functions like Channels, Personal Center and Settings are also essential. Including these secondary points allows a more comprehensive evaluation of the GUI agent. 
More details are provided in Appendix \ref{appendix_functional}.

\subsubsection{Long-Horizon Subset}
The Long-Horizon subset tests whether an agent can maintain a high-level goal, correctly decompose it into subtasks, and execute them sequentially without losing track. 
Each task requires at least 20 steps.
As shown in Figures \ref{figure7_1} and \ref{figure7_2}, tasks such as instant delivery (filling in multiple recipient and sender fields) or adjusting font size (repeatedly clicking the reduce button) evaluate the agent’s ability to manage, sequence, and precisely control subtasks while tracking their completion status.

\begin{table*}[!t]
\setlength{\abovecaptionskip}{0.2cm}
\setlength{\belowcaptionskip}{-0.1cm}
    \centering
\resizebox{1\textwidth}{!}{
    \begin{tabular}{l|l}
    \hline\hline
    Tasks & Task Success Conditions \\  \hline
    Open Bilibili and change my     & \texttt{\!\!1.//*[(@text="Avatar" or @text=“Change Avatar“) and bbox\_contains\_point(../@bounds, \$point)]}  \\
   profile avatar to a random one.\!\!  &  \texttt{\!\!2.//*[@text="Shuffle"and bbox\_contains\_point(../@bounds,\$point) and contains(@package,"bili")]\!\!\!\!\!\!} \\ \hline
    \multirow{3}{*}{\makecell[l]{ Find the subway route from \\ South Railway Station to\!\!\\   Fengtai Station  on Amap.\\
   }}  & \texttt{\multirow{3}{*}{\makecell[l]{\!\!//*[contains(@text,"Public Transportation") and @selected="true" and contains(@package,"map")]\!\!\!\!\!\!\\\!\!and //*[contains(@text,"South Railway Station")  and contains(@resource-id, "summary\_start")]\!\!\!\!\!\!\\ \!\!and //*[contains(@text,"Fengtai Station") and  contains(@resource-id,"summary\_end")].\!\!\!\!\!\!\\}} }\\  \\  \\\hline\hline
    \end{tabular}}
    \caption{Example tasks and their corresponding success
conditions.}
    \label{table_3}
\end{table*}

\subsubsection{GUI-Reasoning Subset}

The GUI-Reasoning subset replicates real-world scenarios requiring active exploration and complex reasoning. Instead of just interacting with obvious elements, tasks involve interpreting visual context, learning through trial and error, and recovering from dead ends. These tasks assess three advanced exploration abilities, as shown in Figure \ref{figure3_2}:

$\bullet$  Icon Understanding: Interpreting visual elements without text labels (weight: 0.5).

$\bullet$ Hidden Function Discovery: Finding non-visible features, like “Draft” and "Settings" hidden in a submenu or an off-screen area (weight: 1).

$\bullet$  Hierarchical Navigation: Efficiently navigating deep menu structures to find a specific function, e.g., “Auto Downloads” via “Settings → Download Settings → Auto Downloads” (weight:2). Hidden Function Discovery is page-level, while Hierarchical Navigation involves app-level menu navigation.

\textbf{Difficulty Levels Based on Exploration Weight:}
Each exploration ability has an assigned weight. Task difficulty is calculated by summing the required ability weights: Easy (weight$\leq$1), Medium (1<weight$\leq$2), and Hard (weight>2).

Table \ref{table_5} shows several examples. 
An easy task, like navigating to a scan function (weight: 0.5), requires only recognizing the Scan icon.
A hard task like contacting customer service (weight: 3) has fewer golden steps but requires more complex exploration, including understanding the app layout and scrolling to find the customer service function. 
Similarly, setting a timer (weight: 4) demands navigating hierarchical menus and locating two hidden functions, Settings and Timer.


\subsubsection{Noise-Robust Subset}
In real-world environments, app actions can lead to unexpected GUI states. The Noise-Robust subset tests an agent’s ability to autonomously recover from disruptions. Following \citet{yang2025gui}, we categorize noise into four types (see Figure \ref{figure3_2}):

$\bullet$ Repeat: The action executes multiple times, often due to insufficient waiting time after execution.

$\bullet$ Unexecuted: The agent generates an action, but the system does not execute this action.

$\bullet$ Delay: The GUI page remains in a prolonged loading state, blocking access to updated elements.

$\bullet$ Pop-up: An unexpected dialog interrupts the workflow  (e.g., a permission dialog or an ad).

\noindent
\textbf{Noise Injection:}
We use Base Subset tasks as seeds and simulate noise during trajectory generation. 
At each step, one of four noise types occurs with a 20\% probability.
For Repeat and Unexecuted noise, the action is either repeated or skipped.
For Delay and Pop-up noise, we predefine five GUI pages per app and randomly select one to display.
For Delay noise, if the agent acts instead of waiting, the action executes on the next page. 
For Pop-up noise, the GUI remains blocked by the pop-up until the correct close element is clicked. 
Further details are provided in Appendix \ref{appendix_noise_robust}.

\begin{table}[!t]
\setlength{\abovecaptionskip}{0.2cm}
\setlength{\belowcaptionskip}{-0.1cm}
    \centering
\resizebox{1\columnwidth}{!}{
    \begin{tabular}{c|c|c|c|c|c}
    \hline\hline
    \multirow{2}{*}{Subset}  & \multirow{2}{*}{Base} & \!\! Long-\!\! & Long- &  \!\!GUI-\!\! & Noise- \\ 
     & & \!\! Tail\!\! & Horizon &  \!\!Reasoning\!\! & Robust \\  \hline
    Tasks &  310 & 340 & 60 & 60 &  310\\ \hline
    Avg Steps &  5.61 & 5.38 & 22.73 & 6.22 &  5.61 \\ \hline
    \hline
     Easy &  196 & 269 &  -  & 13 & 196 \\ \hline
     Medium &  114 & 80 &  -  & 24 & 114 \\\hline
     Hard &  - & - &  60  & 23 & -\\ \hline\hline
    Tasks Per App&  20-30 & 5 & 5 & 5 &  20-30 \\\hline
    Apps   &  12 & 68 & 12 & 12 & 12 \\\hline
      \!\!Apps Per Category\!\!&  1 & 5-6 & 1 & 1 & 1 \\ \hline
    \!\!Functional Points\!\! &  28 & 11 & 8 & 18  &  28 \\\hline\hline
    \end{tabular}}
    \caption{Dataset statistics of MobileBench-OL.}
    \label{table_4}
\end{table}

\begin{figure}[!tb]
\setlength{\abovecaptionskip}{0.2cm}
\setlength{\belowcaptionskip}{-0.2cm}
\centering
\includegraphics[width=0.93\columnwidth]{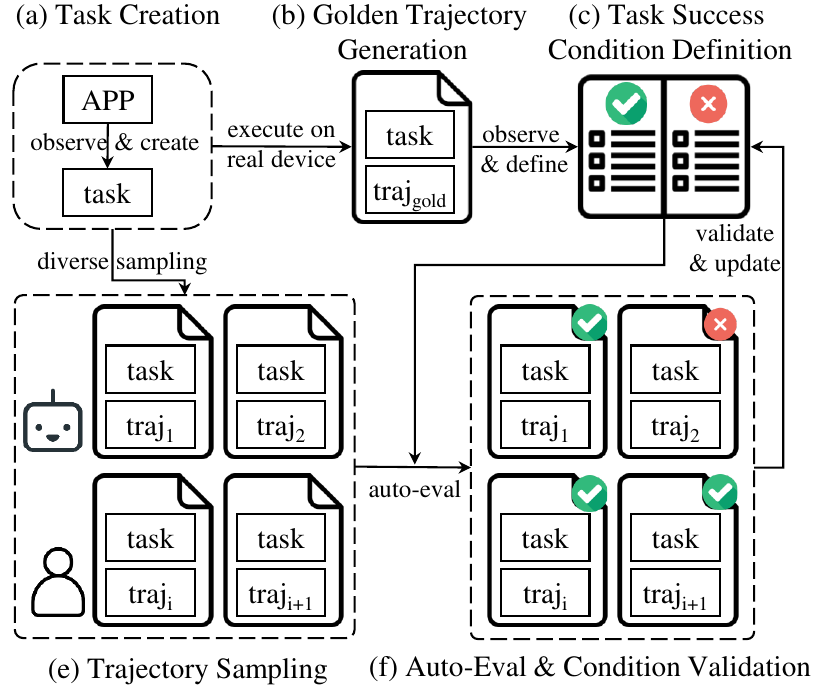} 
\caption{Data construction pipeline.}
\label{figure8_1}
\end{figure}

\begin{figure*}[tbh]
\setlength{\abovecaptionskip}{0.3cm}
\setlength{\belowcaptionskip}{-0.1cm}
\centering
\includegraphics[width=0.96\linewidth]{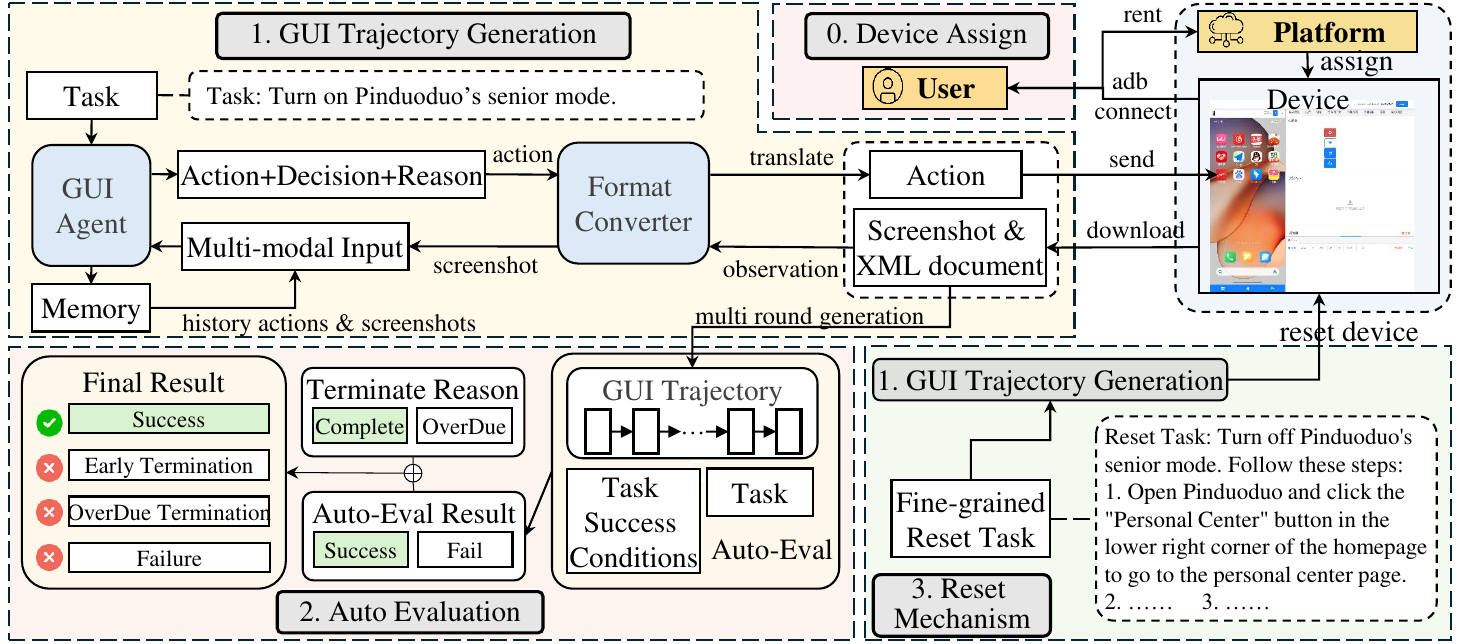}
\caption{Pipeline of the Auto-Eval framework.}
\label{figure_5_1}
\end{figure*}

\subsection{Data Construction}\label{section_condition}
Table \ref{table_4} shows the statistics for MobileBench-OL. See Appendix \ref{appendix_dataset_statistics} for more details.

\textbf{Data construction pipeline:}
Tasks are manually created from scratch to avoid LLM data contamination. The pipeline (Figure \ref{figure8_1}) includes five steps: (1) Annotators design and run tasks on real devices to obtain golden trajectories. (2) Domain experts define task success conditions based on the task-trajectory pairs. (3) UI-TARS-1.5 and  annotators sample multiple trajectories for diverse completion scenarios. (4) An auto-eval framework assesses trajectories against the success conditions. (5) Experts verify the auto-eval results, updating the task success conditions based on any discrepancies. Each task has $\geq$5 samples for reliability.

\textbf{Annotation of Task Success Conditions:}
Task success in MobileBench-OL is determined by rule-based conditions using UI elements and actions observed only in successful trajectories. 
These conditions ensure: (1) \textit{Completeness}: every successful trajectory meets the condition; (2) \textit{Soundness}: no failed trajectory fully meets it. 
This design can match various valid trajectories. 
For example, task "Change to a random Bilibili avatar" is satisfied simply by clicking the "Change Avatar" and then the "Shuffle", regardless of the navigation path.
 
\textbf{Standardized Implementation:}
For scalability, conditions are uniformly defined using XPath-like rules based on GUI attributes like text, resource-id, and package (see Appendix \ref{appendix_condition}). 
In Table \ref{table_3}, one sub-condition uses \texttt{contains(@package,"bili")}
and \texttt{text="Avatar"} to locate the Avatar button, while \texttt{bbox\_contains\_point} confirms the interaction occurred within its bounds. Another sub-condition locates an element with \texttt{text="Shuffle"}. Combining these sub-conditions allows for reliable identification of successful trajectories.

\section{Auto-Eval Pipeline}

\noindent
\textbf{Auto-Eval Framework: }
The Auto-Eval framework consists of GUI Trajectory Generation and Auto Evaluation,
as shown in Figure \ref{figure_5_1}. The single-step trajectory loop iteratively (1) observes the device state and interaction history to form multimodal input, (2) generates actions from this input, (3) translates actions into a unified format and executes them to update the device. This loop repeats until the agent triggers a complete action or the trajectory reaches the max length limit.

Once the trajectory ends, the framework evaluates it against the task success conditions and generates an auto-eval result:
A trajectory meeting all conditions and ending with a "Complete" action is marked Success.
If it meets all conditions but reaches the step limit, it is Overdue Termination.
If it ends with a "Complete" action but does not meet all conditions, it is Early Termination.
Otherwise, the trajectory is marked Failure.

\begin{table*}[!t]
\setlength{\abovecaptionskip}{0.3cm}
\setlength{\belowcaptionskip}{-0.1cm}
    \centering
\resizebox{1\textwidth}{!}{
    \begin{tabular}{l|cc|cc|cc|cc|cc}
    \hline\hline
        \multirow{2}{*}{Model}   &   \multicolumn{2}{c|}{Base}  & \multicolumn{2}{c|}{Long-Tail} & \multicolumn{2}{c}{Long-Horizon} & \multicolumn{2}{c}{GUI-Reasoning}  & \multicolumn{2}{c}{Noise-Robust}   \\ \cline{2-11} 
        & SR ($\uparrow$)  & Sub SR ($\uparrow$) &  SR ($\uparrow$)  & Sub SR ($\uparrow$) & SR ($\uparrow$)  & Sub SR ($\uparrow$)  & SR ($\uparrow$)  & Sub SR ($\uparrow$) & SR ($\uparrow$)  & Sub SR ($\uparrow$)   \\ \hline 
        GPT-4o & 30.32\%& 34.75\%& 22.35\%& 24.95\%& 0.00\%& 7.27\%& 21.67\%& 25.67\%& 27.10\%& 31.10\%\\
        T3A  &  21.61\%& 23.74\%&  8.24\%& 9.56\%& 1.67\%& 9.84\%&  13.33\%& 16.67\%& 5.81\%& 9.87\%\\ 
        M3A  &  40.65\%& 43.88\%&  25.88\%& 28.09\%& 3.33\%& 18.12\%&  25.00\%& 33.06\%& 16.45\% & 23.47\%\\ 
        Mobile-Agent-V2  &  46.45\%& 48.66\%&  33.53\%& 35.88\%& 3.33\% & 19.26\%&  41.67\%& 44.17\%& 37.42\%& 38.23\%\\  \hline
        InternVL2-8B & 0.32\%  & 0.59\%  & 0.29\%  &   0.92\%  & 0.00\%& 0.94\%& 0.00\% & 0.00\% & 0.00\%& 0.43\%\\
        CogAgent-9B  & 4.19\%&   4.30\%  & 1.47\%  & 2.21\%  & 0.00\%& 2.25\%& 0.00\% & 0.00\%& 2.26\%& 2.53\%\\
        UGround-V1-7B   & 0.97\% &   1.77\% & 0.88\%    & 2.12\%  & 0.00\%& 3.44\%& 0.00\% & 0.83\% & 0.00\%& 1.41\%\\
        OS-Atlas-Pro-7B    & 2.58\%  & 13.33\% & 0.00\%  &   11.76\% & 0.00\%& 8.64\%& 1.67\% & 7.83\% & 0.65\%& 8.76\%\\
        Qwen2-VL-7B & 10.32\%&  19.46\%& 5.29\%  &  17.28\% & 0.00\%& 9.50\%& 0.00\% & 1.67\% & 4.84\%& 16.24\%\\
        Qwen2.5-VL-7B  & 31.61\%& 37.63\%& 25.88\% &   27.65\% & 0.00\%& 16.30\%& 16.67\% & 23.44\% & 30.97\%& 35.01\%\\
        UI-TARS-7B    & 46.77\% &  57.06\% & 32.94\%  & 41.32\% & 1.67\% & 19.74\%& 23.33\%& 33.17\%& 40.97\% & 49.06\%\\
        UI-TARS-1.5-7B & \textbf{60.97\%} &  \textbf{64.84\%} & \textbf{41.76\%}  & \textbf{46.32\%} &  \textbf{15.00\%} & \textbf{45.25\%}&\textbf{38.33\%}& \textbf{40.83\%}& \textbf{56.77\%} & \textbf{62.18\%}\\\hline\hline
    \end{tabular} }
    \caption{The overall performance of GUI agents on MobileBench-OL. 
    }
    \label{result_1}
\end{table*}

\textbf{Reset Mechanism: }
In real-world environments, device states constantly change during task execution, and some tasks cause persistent changes that can affect later evaluations. For example, “Open Senior Mode” changes the UI style permanently, while "Setting Company Address" allows users to skip future reconfigurations after one successful set. To address this, we propose a reset mechanism that uses  a reliable agent to execute fine-grained inverse tasks. As shown in Figure \ref{figure_5_1}, the inverse task of “Turn on Pinduoduo’s senior mode” is defined as a step-by-step fine-grained description of how to "turn off Pinduoduo’s senior mode", including the shape and position of interactive elements.

Reset tasks fall into four categories:
(1) Task-level reset: Adjusting app state or user data for a single task. (2) App-level reset: A shared reset (e.g., Clear Shopping Cart) applicable to multiple tasks in an app. (3) No reset needed: Tasks that only require viewing or navigation. (4) Infeasible: Tasks with server interaction or irreversible actions.
For infeasible tasks, we expand task success conditions to ensure reliable evaluation. For example, a sign‑in task succeeds if either the “Sign In” button is clicked or a “Signed in today” element appears.

After each round of benchmark evaluation, 255 task-level and app-level reset tasks are executed to restore the device state.
Experiments have shown that the reset mechanism has a success rate exceeding 90\%. More details refer to Appendix \ref{appendix_reset}.




\section{Experiment}

\subsection{Experiments Setting}

We define four key metrics for agent performance: (1) Success Rate (SR): The primary metric, indicating a task is successfully completed and ends with a "Complete" action.
(2) Sub-condition Success Rate (Sub-SR): Measures the ratio of completed sub-conditions, regardless of how the task ended.
(3) Step Ratio: Compares the agent's steps to the ideal human golden steps, with a max limit of three times the golden steps.
(4) Failure Reasons: Categorized into three types: Early Termination, Overdue Termination, and Failure. 

We test 12 GUI agents: 4 closed-source (GPT-4o \cite{yan2023gpt4vwonderlandlargemultimodal}, M3A \cite{rawles2024androidworld}, T3A \cite{rawles2024androidworld}, Mobile-Agent-V2 \cite{wang2024mobile}) and 8 open-source (InternVL2 \cite{wang2024mpo}, CogAgent \cite{hong2023cogagent}, UGround-V1  \cite{gou2025uground}, OS-Atlas \cite{wu2024atlas}, Qwen2-VL  \cite{Qwen2-VL}, Qwen2.5-VL \cite{Qwen2.5-VL}, UI-TARS  \cite{qin2025ui}, and UI-TARS-1.5  \cite{qin2025ui}).
All open-source models are of comparable capacities(7B,8B,9B). 
The step limit is set to three times the golden steps, with a 3-second wait after each action. Details of  metrics and implementation refer to Appendix \ref{appendix_experiments}.

\begin{table}[!t]
\setlength{\abovecaptionskip}{0.3cm}
\setlength{\belowcaptionskip}{-0.1cm}
    \centering
\resizebox{1\columnwidth}{!}
{
    \begin{tabular}{l|ccc}
    \hline\hline
         Long-Horizon    & Total Tasks & Failed Tasks \\ \hline 
         SR Tasks  	& 15.0\%	& -- \\ \hline
         Reasoning \& Planning Failure & 15.0\% &	17.6\% \\ 
         Subtask Omission & 6.7\% & 7.8\% \\ 
         Function Navigation Failure & 16.7\% & 19.6\% \\ 
         Attribute Omission/Error & 25.0\% & 29.4\% \\ 
         Visual Grounding Failure & 21.7\% & 25.5\% \\  \hline\hline
    \end{tabular}}
    \caption{Error Distribution of UI-TARS-1.5 across 60 tasks in Long-Horizon Subset.
    }
    \label{result_8}
\end{table}

\begin{table}[!t]
\setlength{\abovecaptionskip}{0.2cm}
\setlength{\belowcaptionskip}{-0.1cm}
    \centering
\resizebox{0.85\columnwidth}{!}
{
    \begin{tabular}{l|ccc}
    \hline\hline
        GUI-Reasoning  & Easy & Medium & Hard \\\hline 
         M3A & 38.46\%&37.50\%& 4.35\%\\ 
         T3A & 23.08\%&16.67\%& 4.35\%\\ 
         Mobile-Agent-V2 & 53.85\%&50.00\%& 26.09\%\\ \hline
         Qwen2.5-VL  & 30.77\%	& 	16.67\%	& 	8.70\%	\\ 
         UI-TARS & 30.77\%& 	25.00\%	& 	17.39\%	 \\ 
         UI-TARS-1.5 & 61.54\%	& 	41.67\%	& 	21.74\%	 \\ \hline\hline
    \end{tabular}}
    \caption{Success Rate across GUI-Reasoning difficulty.
    }
    \label{result_9}
\end{table}

\begin{table}[!t]
\setlength{\abovecaptionskip}{0.2cm}
\setlength{\belowcaptionskip}{-0.1cm}
    \centering
\resizebox{1\columnwidth}{!}
{
    \begin{tabular}{l|cccc}
    \hline\hline
     Noise-Robust  &  Repeat & Unexecuted & Delay & Pop-Up  \\ \hline 
         M3A &  22.78\% & 16.67\%& 22.37\%& 3.90\%  \\ 
T3A & 10.13\% & 3.85\%& 3.95\% & 5.19\%\\ 
Mobile-Agent-V2  & 49.37\%& 43.59\% & 40.79\%& 15.58\%  \\ \hline 
         Qwen2.5-VL & 32.91\%& 37.18\%&  25.00\%& 28.57\%  \\ 
         UI-TARS & 43.04\%& 37.18\%& 46.05\%& 37.66\%\\ 
         UI-TARS-1.5 & 60.76\%& 60.26\%& 60.53\%& 41.56\% \\ \hline\hline
    \end{tabular}}
    \caption{Success Rate across different noise types.
    }
    \label{result_7}
\end{table}

\begin{table*}[!t]
\setlength{\abovecaptionskip}{0.3cm}
\setlength{\belowcaptionskip}{-0.1cm}
    \centering
\resizebox{1\textwidth}{!}
{
    \begin{tabular}{l|ccc|ccc|ccc|ccc|ccc}
    \hline\hline
         & \multicolumn{3}{c|}{Base}  & \multicolumn{3}{c|}{Long-Tail} & \multicolumn{3}{c|}{Long-Horizon} & \multicolumn{3}{c|}{GUI-Reasoning}  & \multicolumn{3}{c}{Noise-Robust} \\ 
         & Early & Overdue & Failure  & Early & Overdue & Failure  & Early & Overdue & Failure  & Early & Overdue & Failure  & Early & Overdue & Failure \\ \hline 
         M3A & 31.29\% & 1.61\% & 26.45\% & 41.76\% & 0.59\% & 31.76\% & 50.00\% & 0.00\% & 46.67\% & 35.00\% & 5.00\% & 35.00\% & 14.84\% & 5.48\% & 63.23\% \\
         T3A & 73.87\% & 0.00\% & 4.52\% & 86.76\% & 0.00\% & 5.00\% & 95.00\% & 0.00\% & 3.33\% & 85.00\% & 0.00\% & 1.67\% & 37.74\% & 2.90\% & 53.55\% \\
         Mobile-Agent-V2 & 53.55\% & 0.00\% & 0.00\% & 66.47\% & 0.00\% & 0.00\% & 96.67\% & 0.00\% & 0.00\% & 58.33\% & 0.00\% & 0.00\% & 62.58\% & 0.00\% & 0.00\% \\ \hline
        
         Qwen2.5-VL & 39.68\% & 2.90\% & 25.81\% & 31.18\% & 0.59\% & 42.35\% & 28.33\% & 0.00\% & 71.67\% & 45.00\% & 3.33\% & 35.00\% & 40.00\% & 1.94\% & 27.10\% \\
         UI-TARS & 18.39\% & 7.42\% & 27.42\% & 25.59\% & 6.47\% & 35.00\% & 13.33\% & 3.33\% & 81.67\% & 23.33\% & 5.00\% & 48.33\% & 21.29\% & 6.77\% & 30.97\% \\
         
         UI-TARS-1.5 & 17.42\% & 2.26\% & 19.35\% & 21.76\% & 2.65\% & 33.82\% & 25.00\% & 3.33\% & 56.67\% & 21.67\% & 0.00\% & 40.00\% & 19.35\% & 3.55\% & 20.32\% \\ \hline\hline
    \end{tabular}}
    \caption{Statistics of failure reasons (Early Termination/Overdue Termination/Failure) on all 1080 tasks.
    }
    \label{result_2}
\end{table*}

\subsection{Main Results}

Table \ref{result_1} summarizes the performance of all GUI Agents on MobileBench-OL. We can see that:

$\bullet$  UI-TARS-1.5-7B achieved the highest SR and Sub-SR across all five subsets.

$\bullet$  Performance is the lowest in the Long-Horizon subset due to its comprehensive assessment of task interpretation, decomposition, multi-step execution, and long-term memory.

$\bullet$  Results drops in Long-Tail and GUI-Reasoning subsets because agents struggle with uncommon app structures and lack strong exploration abilities.

$\bullet$ Performance also drops in the Noise-Robust subset, as most agents are unfamiliar with handling interface noise, such as unexpected pop-ups.

$\bullet$ Agents like UGround and InternVL2 have very low SR (<5\%), likely due to poor grounding. They can generate appropriate intents during execution, but often fail to produce the correct actions.

$\bullet$ Agents that accept multiple images as input (e.g., GPT-4o) achieve higher SR. This allows them to observe state changes caused by previous actions, thus aiding in error correction and loop recovery.

$\bullet$ Qwen2-VL-7B shows a large gap between its SR and Sub-SR because it often fails to understand when to terminate a task. 

$\bullet$ The Sub-SR metric closely aligns with the SR in most subsets except Long-Horizon. This is because Long-Horizon tasks are complex and have many completion conditions, making it difficult for the agent to finish the entire task even when some sub-tasks are completed.

\subsection{Ablation Study}
\textbf{Long-Horizon Error Analysis}
Table \ref{result_8} classifies errors in the Long-Horizon subset into four categories.
Attribute errors (25.0\%)) were most frequent, showing difficulty in tracking attributes over long-term trajectories.
The other three categories, which relate to the agent's grounding, task intent understanding, and function navigation abilities, also need improvement.
Subtask omissions (6.7\%) were least common but highlight the risk of missing subtasks steps in complex tasks.
See Appendix \ref{appendix_long_horizon_error} for detailed definitions and examples.

\noindent
\textbf{GUI-Reasoning Difficulty Analysis}  Table \ref{result_9} shows that SR decreases as exploration difficulty increases in GUI-Reasoning subset. 
The agent performs well on Easy tasks, which only require simple icon recognition or finding one hidden function. Performance drops sharply on Hard tasks, which require locating multiple hidden functions or deeper navigation.

\noindent
\textbf{Noise Type Analysis}
Table \ref{result_7} shows that Pop-Up noise has the strongest negative effect in the Noise-Robust subset. While other noises mainly test the ability to handle unexpected action results or recover from an error state, Pop-Up noise specifically requires understanding pop-up windows and accurately locating the close icon. Many agents fail to close the pop-up window, causing them to become stuck on the pop-up page, terminating the task early, or exceeding step limits.

\begin{table}[!t]
\setlength{\abovecaptionskip}{0.2cm}
\setlength{\belowcaptionskip}{-0.1cm}
    \centering
\resizebox{\columnwidth}{!}
{
    \begin{tabular}{l|cc|cc|c|cc }
    \hline\hline 
          & \multicolumn{2}{c|}{Base}  & \multicolumn{2}{c|}{Long-Tail} & Long-Horizon & \multicolumn{2}{c}{Noise-Robust}  \\ 
         & Easy & Medium & Easy & Medium & Hard  & Easy & Medium   \\\hline 
         M3A & 46.94\%&29.82\%& 28.08\%& 18.75\%&-& 19.90\%& 10.53\%\\ 
         T3A & 26.53\%&13.16\%& 9.23\%& 5.00\%&-	& 8.16\%& 1.75\%\\ 
         Mobile-Agent-V2 & 53.06\%&35.09\%& 37.31\%& 21.25\%&-	& 43.88\%& 26.32\%\\ \hline
         Qwen2.5-VL & 42.35\%& 13.16\%& 31.54\% & 7.50\%  & -	& 41.84\%& 12.28\%\\ 
         UI-TARS & 52.55\% & 36.84\%  & 36.15\%  & 22.50\% & 1.67\%  & 44.39\%& 35.09\%\\ 
         UI-TARS-1.5 & 65.31\%  & 53.51\% & 46.15\% &27.50\% & 15.00\%  & 62.76\%& 46.49\%\\ \hline\hline
    \end{tabular}}
    \caption{SR across different Length Difficulty Levels.
    }
    \label{result_5}
\end{table}

\begin{table}[!t]
\setlength{\abovecaptionskip}{0.3cm}
\setlength{\belowcaptionskip}{-0.1cm}
    \centering
\resizebox{1\columnwidth}{!}
{
    \begin{tabular}{l|cccc|cc|c}
    \hline\hline
         & \multicolumn{4}{c|}{Num}  & \multicolumn{2}{c|}{Success Rate}  & Auto-Eval \\\cline{2-7} 
         & TP & FP & FN & TN & Auto-Eval & Human & Correct \\\hline 
         Base & 185  &3 &12  &110 & 60.97\% & 63.54\% & 95.16\%  \\
         Long-Tail &145&0  & 5&  190 &41.76\% & 44.12\%& 98.53\% \\ 
         Long-Horizon & 9 & 0 & 0 & 51 & 15.00\% & 15.00\% & 100.00\%  \\
         GUI-Reasoning & 22 & 1 & 2 & 35 & 38.33\% & 40.00\% & 95.00\% \\
         Noise-Robust & 173 & 1 & 3 & 133 & 56.77\% & 56.77\% & 98.71\% \\ \hline
         Overall & 534 & 5 & 22 & 519 & 49.91\% & 51.48\% & 97.50\%  \\\hline\hline
    \end{tabular}}
    \caption{Human evaluation of Auto-Eval results.
    TP/FP/FN/TN follow the confusion matrix definition.  
    Last column shows Auto-Eval's accuracy relative to human judgment.
    }
    \label{result_3}
\end{table}

\begin{table}[!t]
\setlength{\abovecaptionskip}{0.2cm}
\setlength{\belowcaptionskip}{-0.1cm}
    \centering
\resizebox{1\columnwidth}{!}
{
    \begin{tabular}{l|c|c|c|c}
    \hline\hline
          &  Base  & Long-Tail & \!\!\!\!GUI-Reasoning\!\!\!\! & Noise-Robust \\  \hline 
         Reset Tasks & 65 & 102 & 23 & 65 \\\hline 
         Auto-Eval SR &95.38\%   & 92.86\%  &91.30\%  &95.38\% \\ 
         Human  Eval SR & 96.92\%  &94.35\% &95.65\% &92.31\%\\ \hline\hline
    \end{tabular}}
    \caption{Human evaluation of Reset Mechanism.
    }
    \label{result_4}
\end{table}

\noindent
\textbf{Failure Reasons}
Table \ref{result_2} summarizes failure reasons for top-performing GUI agents.
Failure is most common, followed by Early Termination, where the agent misinterprets the task or stops prematurely. 
Overdue Termination occurs less frequent, typically when the agent fails to recognize task completion and enters an action loop.
Closed-source agents like Mobile-Agent-V2 have a higher rate of Early Termination, indicating a more aggressive stopping threshold and higher confidence.

\noindent
\textbf{Difficulty Analysis}
Table \ref{result_5} shows SR across difficulty levels based on golden steps. Performance declines with increasing task length in all four subsets, especially in the Long-Horizon tasks, as the agent struggled to complete these complex multi-subtask scenarios.
Overall, MobileBench-OL presents a considerable challenge and leaves room for further improvement.

\noindent
\textbf{Human evaluation of Auto-Eval Framework}
To verify the effectiveness of Auto-Eval Framework, we conducted a human evaluation of UI-TARS-1.5 results, as shown in Table \ref{result_3}. False Negatives (FN) show it sometimes classifies correct GUI trajectories as fail, while False Positives (FP) confirm that Auto-Eval rarely mislabels incorrect trajectories as successful, proving its soundness. Overall, Auto-Eval achieves more than 95\% accuracy, indicating that it can reliably label GUI trajectories.

\noindent
\textbf{Human evaluation of Reset Mechanism}
We also verified the Reset Mechanism's effectiveness via human evaluation. In Table \ref{result_4}, we executed 255 fine-grained reset tasks with UI-TARS-1.5 and evaluated the result using both Auto-Eval and human evaluation. Human evaluation show an accuracy of more than 90\%, indicating the effectiveness of reset mechanism. In addition, Auto-Eval results is close to human evaluations, further verifying the reliability of Auto-Eval.


\section{Conclusion}

This paper introduces MobileBench-OL, an online benchmark for evaluating GUI agents in real-world environments. It contains 1080 tasks from 80 Chinese apps, integrating 5 subsets that measure not only task execution but also complex reasoning, exploration ability, and robustness to noise. To ensure stable and reproducible evaluation, we provide an auto-eval framework with a remote mobile control platform and reset mechanism. Experimental results evaluating 12 leading agents show significant room for improvement to meet real-world requirements. Human evaluation further confirms the benchmark's stability and reliability.

MobileBench-OL assesses multiple critical dimensions for real-world agent performance. We hope that MobileBench-OL can serve as a real, fair and scalable benchmark, and contribute to future research in the community.

\section*{Limitations}



While MobileBench-OL offers a comprehensive online benchmark, it still has some limitations that may be addressed in future updates. While the benchmark already includes long-horizon tasks requiring multiple subtasks and GUI-reasoning tasks requiring exploration, it currently lacks complex cross-app tasks. These would require the combined use of multiple apps to complete several subtasks. We plan to address these more complex scenarios in future updates.

\section*{Ethics Statement}


This paper is conducted in accordance with the ACM Code of Ethics. 
We strictly filtered the benchmark, removing any data that could potentially expose personal privacy, thereby ensuring the highest level of protection for personal data. All data labeling was completed by crowdsourced workers, whom we paid at least \$0.5 for each step and provided with necessary training. The human evaluation of our work was carefully conducted by IT experts. We ensured the reviewers had gender balance and diverse educational backgrounds, reflecting a wide range of perspectives and experiences.


\bibliography{acl_latex}

\clearpage
\appendix

\begin{table*}[htb]
    \centering
\resizebox{1\textwidth}{!}{
    \begin{tabular}{l|l|ll|l|l|l}
    \hline\hline
    \multicolumn{7}{c}{Base}  \\\hline
    App Name	&Category	&APK & & App Name	&Category	&APK \\\hline
BiliBili	&Video	&bili\_8.39.0.apk & & NeteaseMusic	&Music	&netease\_7.23.1.apk\\
ArticleNews	&News	&articlenews\_11.6.0.apk & & FanqieRead	&Reading	&fanqieread\_6.6.7.32.apk\\
Pinduoduo	&Shopping	&pinduoduo\_7.53.0.apk & & Minimap	&Transportation	&minimap\_15.11.1.2030.apk\\
QQ	&Social	&qq\_9.1.91.apk & & Wuba	&Lifestyle	&wuba\_13.29.5.apk\\
Tonghuashun	&Finance	&tonghuashun\_11.30.02.apk & &  Baidubrowser	&Tools	&baidubrowser\_15.11.0.10.apk\\
DingDing	&Office	&dingding\_7.6.55.apk & & 
Seeyou	&Sports	&seeyou\_8.87.1.0.apk\\\hline\hline
    \multicolumn{7}{c}{Long-Tail}  \\ \hline
    App Name	&Category	&APK & & App Name	&Category	&APK \\\hline
idxyer	&Sports	&idxyer\_10.3.0.apk & &
keep	&Sports	&keep\_8.5.30.apk\\
xiaoxun	&Sports	&xiaoxun\_1.2.19.apk & &
dwbtime	&Sports	&dwbtime\_11.6.8.apk\\
dongqiudi	&Sports	&dongqiudi\_8.4.2.apk & &
hupu	&Sports	&hupu\_8.0.50.apk\\
kugoumusic	&Music	& kugoumusic\_20.2.2.apk & &
qtradio	&Music	&qtradio\_10.8.7.apk\\
dreamina	&Music	&dreamina\_1.5.8.apk & &
qqmusic	&Music	&qqmuisc\_14.6.0.8.apk\\
migumusic	&Music	&migumusic\_7.47.0.apk & &
theater	&Video	&theater\_2.7.1.2.apk\\
haokan	&Video	&haokan\_7.83.0.10.apk& &
huajiao	&Video	&huajiao\_9.6.4.2030.apk\\
xiguavideo	&Video	&xiguavideo\_9.6.4.apk& &
tencentvideo	&Video	&tencentvideo\_9.01.60.30228.apk\\
yangcong345	&News	&yangcong\_7.87.0.apk& &
huatu	&News	&huatu\_7.4.350.apk\\
xunfeiai	&News	&xunfeiai\_3.0.4.apk& &
jiakao	&News	&jiakao\_8.80.0.apk\\
zuoyebang	&News	&zuoyebang\_14.30.0.apk& &
wukong	&News	&wukong\_13.0.0.apk\\
icredit	&Office	&icredit\_19.2.0.apk& &
workschedule	&Office	&workschedule\_2.0.91.apk\\
netease	&Office	&netease\_7.23.1.apk& &
kdweibo	&Office	&kdweibo\_10.8.12.apk\\
mobilemcloud	&Office	&mobilemcloud\_12.1.0.apk& &
camscanner	&Office	&camscanner\_6.91.5.apk\\
weread	&Reading	&weread\_9.3.4.apk & &
dmzj	&Reading	&dmzj\_3.9.14.apk\\
fcatfreader	&Reading	&fcatfreader\_7.76.apk& &
jinjiang	&Reading	&jinjiang\_6.6.7.apk\\
qidian	&Reading	&qidian\_7.9.417.apk& &
mtxx	&Reading	&mtxx\_11.12.0.apk\\
smart360	&Tools	&smart360\_2.12.1.apk& &
zhejiang	&Tools	&zhejiang\_7.26.0.apk\\
weitdy	&Tools	&weitdy\_1.0.73.apk& &
zhipuai	&Tools	&zhipuai\_3.1.3.apk\\
cbn	&Tools	&cbn\_2.0.2.apk& &
newhope	&Tools	&newhope\_4.0.6.apk\\
vipshop	&Shopping	&vipshop\_9.53.7.apk& &
youpin	&Shopping	&youpin\_5.31.0.apk\\
yonghui	&Shopping	&yonghui\_11.6.0.1.apk& &
vmall	&Shopping	&vmall\_1.25.5.300.apk\\
xiaomishop	&Shopping	&xiaomishop\_5.41.0.apk& &
motorfans	&Transportation	&motorfans\_3.65.00.apk\\
zeekrlife	&Transportation	&zeekrlife\_4.9.11.apk& &
autohome	&Transportation	&autohome\_11.75.5.apk\\
aima	&Transportation	&aima\_5.3.2.apk& &
htinns	&Transportation	&htinns\_9.34.0.apk\\
szzc	&Transportation	&szzc\_9.2.0.apk& &
ivwen	&Social	&ivwen\_11.0.6.apk\\
momo	&Social	&momo\_9.17.6.apk& &
hsj	&Social	&hsj\_2.9.8.apk\\
maimai	&Social	&maimai\_6.6.72.apk& &
douban	&Social	&douban\_7.104.0.apk\\
cc5	&Lifestyle	&cc5\_10.15.0.apk& &
calendar	&Lifestyle	&calendar\_7.2.6.apk\\
moviepro	&Lifestyle	&moviepro\_8.8.4.apk& &
wsgw	&Lifestyle	&wsgw\_3.1.6.apk\\
starbucks	&Lifestyle	&starbucks\_10.11.0.apk& &
xiachufang	&Lifestyle	&xiachufang\_8.8.65.apk\\
lottery	&Finance	&lottery\_3.6.6.apk& &
miaomiao	&Finance	&miaomiao\_4.1.7.apk\\
lifecircle	&Finance	&lifecircle\_2.75.1.apk& &
cailianshe	&Finance	&cailianshe\_8.6.2.apk\\
creditcard	&Finance	&creditcard\_6.2.9.apk& &
gszq	&Finance	&gszq\_9.09.000.apk \\ \hline\hline
    \end{tabular} }
    \caption{The apks of 80 MobileBench-OL Apps.}
    \label{appendix_apps}
\end{table*}

The appendix provides supplementary details to further describe the content of the main section.
Section \ref{appendix_a} introduces MobileBench-OL environment, including the running devices, the apps and APKs used, the defined action space, and the format of the environment data.
Section \ref{appendix_b} details all 5 subset, providing examples and statistics. This includes definitions of metrics such as feature coverage, exploration difficulty, noise classification, and noise injection methods.
Section \ref{appendix_condition} provides an analysis of GUI trajectories and Task Success Conditions.
Section \ref{appendix_reset} introduces the Reset Mechanism.
Section \ref{appendix_g} provides the prompts used for the different baselines.
Section \ref{appendix_related} provides more related works.
Section \ref{appendix_experiments} details the experimental setup, including the Evaluation Metrics, Implementation Details, and Baselines.
Section \ref{appendix_more_experiments} presents ablation experiments not included in the main text.

\section{MobileBench-OL Environment}\label{appendix_a}

\subsection{Real-World Environments}

To ensure the benchmark runs in real-world environments, we use an online mobile control platform. We offer three physical smartphones connected to the platform, each pre-installed with 80 benchmark apps and logged in with real, verified user accounts to ensure authentic app behavior and network conditions. Researchers interact with these devices through our online control platform. The platform provides ADB access; the GUI Agent can send actions to the devices for execution, as well as download environmental data from them, with all operations conducted under real network conditions. In addition, researchers can monitor the device screens in real time through the platform.

\begin{figure}[!t]
\centering
\includegraphics[width=1\columnwidth]{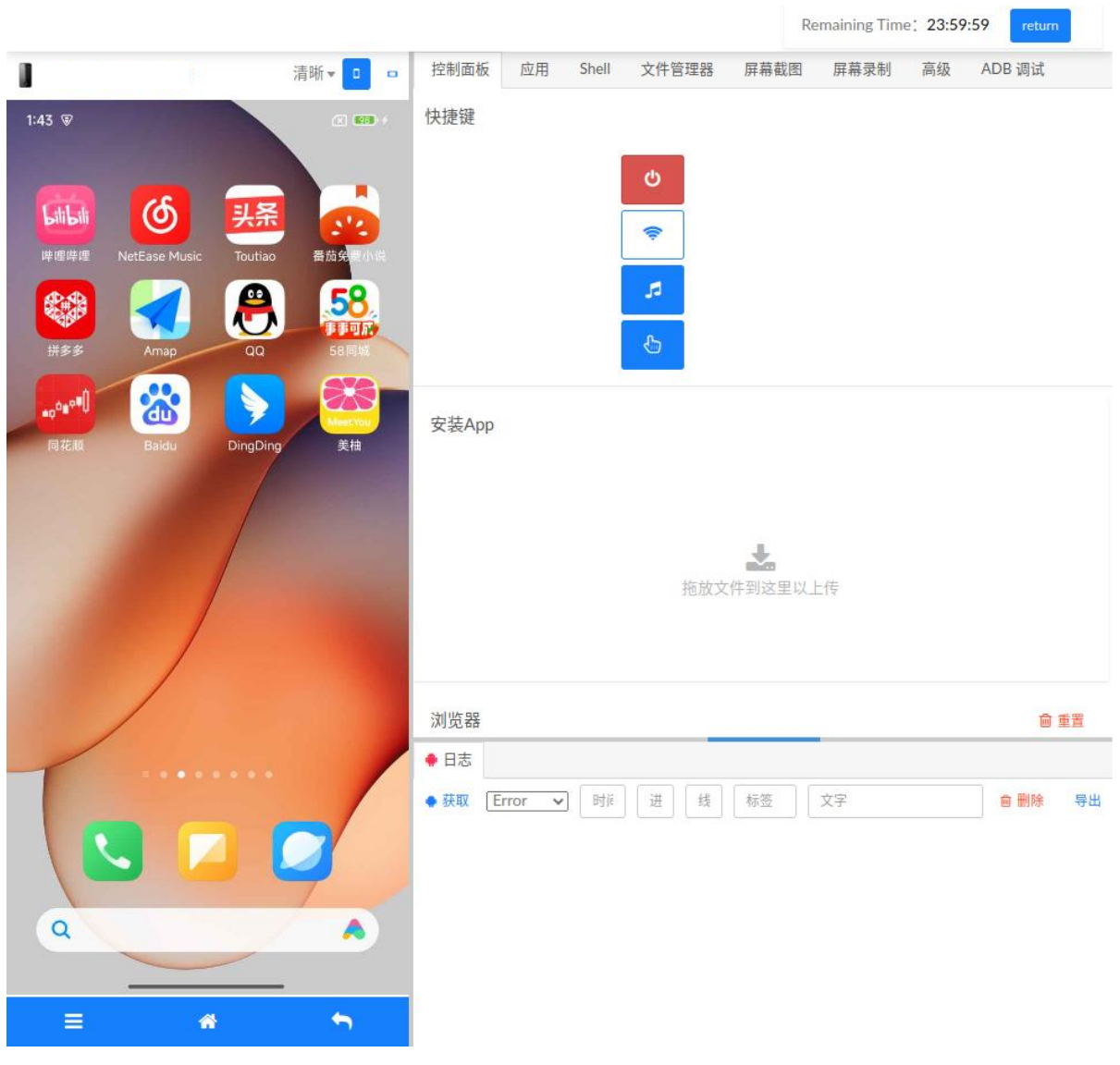} 
\caption{Online platform interface.}
\label{figure_appendix_6_1}
\end{figure}

Figure \ref{figure_appendix_6_1} shows the platform interface. The web interface allows real-time screen viewing, ADB connection via device IP for data download and command execution, and direct screen interaction through clicking and dragging. 

\begin{figure}[!t]
\centering
\includegraphics[width=0.91\columnwidth]{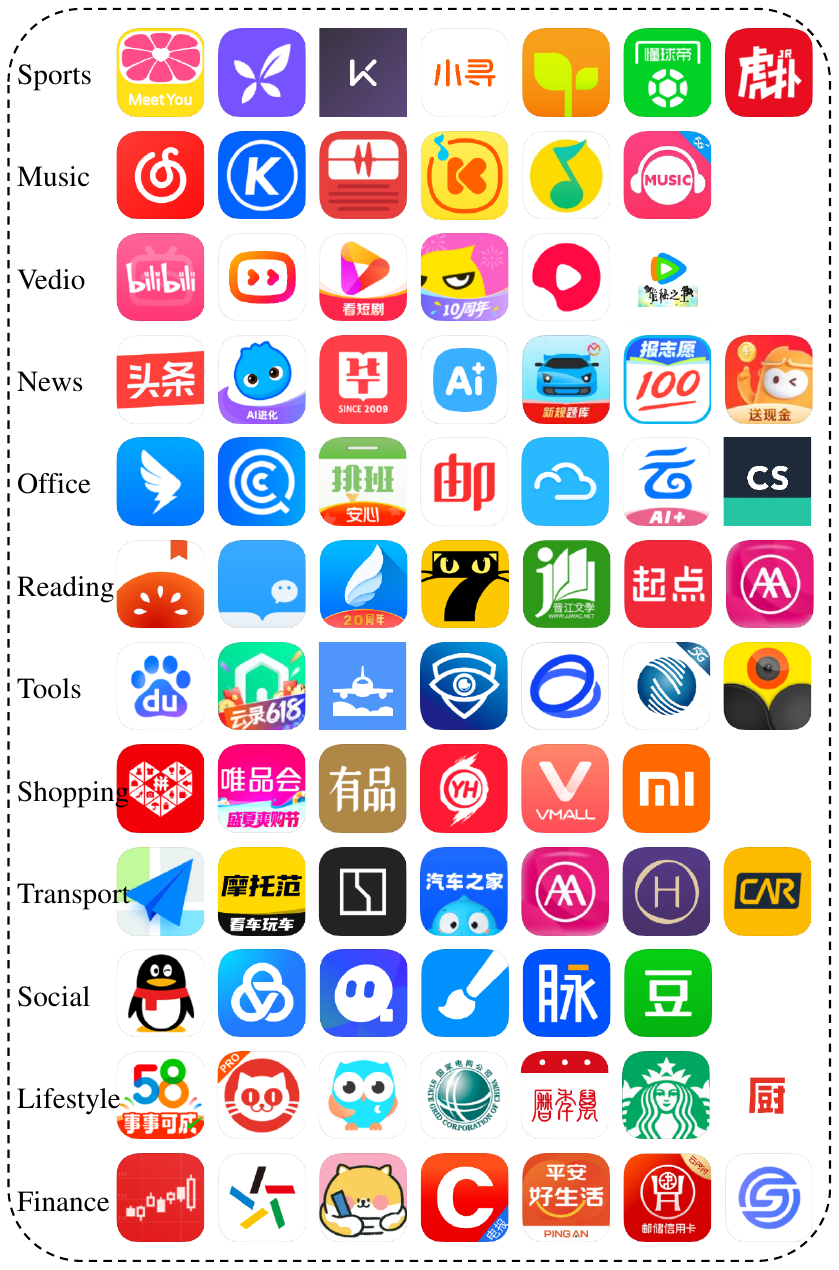} 
\caption{List of all app icons.}
\label{figure_appendix_1}
\end{figure}

Moreover, we will release all APK files, as shown in Figure \ref{figure_appendix_1} and Table \ref{appendix_apps}, allowing users to reproduce the experimental benchmark environment on their own devices. This ensures the benchmark remains accessible beyond the lifecycle of any specific platform. Researchers can simply  run the benchmark on their own mobile phones. All 80 benchmark APK files can be installed via ADB. While our platform uses verified accounts, logging in with any random account during local reproduction does not affect evaluation outcomes. This approach with physical devices is chosen because virtual emulators often fail to run real-world apps stably, support account login, and may crash frequently.


\subsection{APP Selection}
The distribution and categories of 80 apps are presented in Figure \ref{figure_appendix_1}. Table \ref{appendix_apps} listed the APK version for all apps. Researchers can find the corresponding version in the app's version history on the App Store to replicate our test environment. They can also try installing the latest version of the corresponding app. Two months after building our benchmark environment, we tested the latest versions and found that UI changes caused fluctuations in agent performance, but the range of fluctuation did not exceed 5\%. Therefore, if researchers wish only to compare the performance of two agents, they can use the latest version to test both simultaneously. However, to reproduce the exact results in the paper, we recommend using the corresponding version of the APK.

\begin{figure*}[!t]
\centering
\includegraphics[width=1\textwidth]{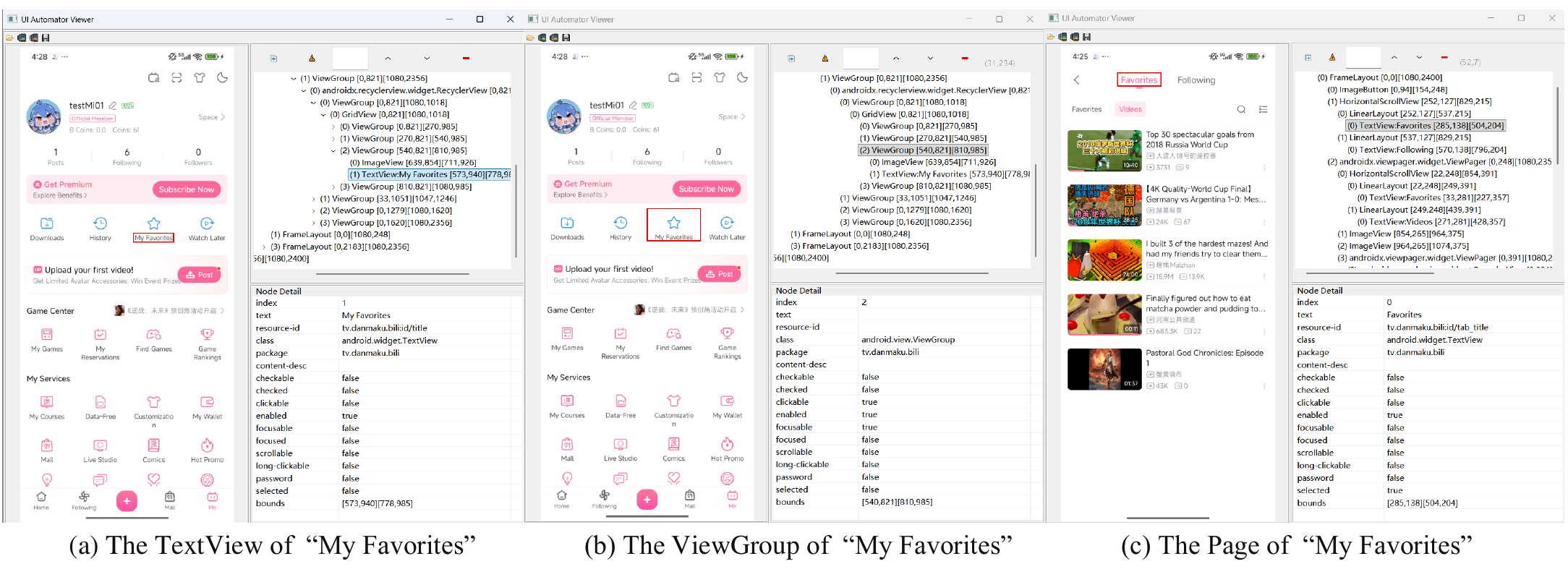} 
\caption{Examples of XML files.}
\label{figure_appendix_6_2}
\end{figure*}

\subsection{Action Space} 
Different GUI agents generate actions in different formats, which we will translate into a unified action format. Our action space includes:

\begin{tcolorbox}[    
    colback=blue!3!white,     
    colframe=blue!40!black,
breakable, 
                  ]
                  
\begin{verbatim}
click(point=’<point>x1 y1</point>’)
type(content=’...’)
scroll(point=’<point>x1 y1</point>’,
        direction=’...’)
press_home()
press_back()
wait()
long_press(point=’<point>x1 y1
        </point>’)
finished(content=’xxx’)
\end{verbatim}
\end{tcolorbox}

\begin{table*}[!t]
    \centering
\resizebox{1\textwidth}{!}{
    \begin{tabular}{c|l|c|c}
    \hline\hline
    \multirow{2}{*}{Subsets} & \multirow{2}{*}{Task Examples} & Steps & \!\!Exploration\!\! \\ 
    &  & Difficulty &  Difficulty \\\hline
    \multirow{10}{*}{Base }   & \multirow{2}{*}{\makecell[l]{ Open Bilibili's animation channel to see what children's \\   animations are available.
   }} &  \multirow{2}{*}{Easy}  & \multirow{2}{*}{-} \\  & & & \\  \cline{2-4}
   & \multirow{2}{*}{\makecell[l]{ On Bilibili, find the updates from the people I follow and \\ like their posts.
   }} &  \multirow{2}{*}{Easy}  & \multirow{2}{*}{-} \\  & & & \\  \cline{2-4}
& \multirow{2}{*}{\makecell[l]{ Check my pending payment orders on Bilibili's \\ membership store.
   }} &  \multirow{2}{*}{Easy}  & \multirow{2}{*}{-} \\  & & & \\  \cline{2-4}
   & \multirow{2}{*}{\makecell[l]{ Find and follow singer Jay Chou's personal homepage \\ on NetEase Cloud Music.
   }} &  \multirow{2}{*}{Medium}  & \multirow{2}{*}{-} \\  & & & \\  \cline{2-4}
   & \multirow{2}{*}{\makecell[l]{ Open NetEase Cloud Music and post a note saying \\  "Good luck with exams."
   }} &  \multirow{2}{*}{Medium}  & \multirow{2}{*}{-} \\  & & & \\  \hline
    \multirow{9}{*}{Long-Tail }   & \multirow{2}{*}{\makecell[l]{ Check on Dingxiang Garden which medication should be \\ used for allergic rhinitis.
   }} &  \multirow{2}{*}{Easy}  & \multirow{2}{*}{-} \\  & & & \\  \cline{2-4}
   & \multirow{2}{*}{\makecell[l]{ Look up the lottery results for traditional football lottery \\ in China Sports Lottery.
   }} &  \multirow{2}{*}{Easy}  & \multirow{2}{*}{-} \\  & & & \\  \cline{2-4}
&  Turn on the "Honk to Find Car" feature in Aima.
    & Easy  & -\\   \cline{2-4}
   & \multirow{2}{*}{\makecell[l]{ Set a shift alarm for the day shift at 19:00 in the \\ scheduling calendar.
   }} &  \multirow{2}{*}{Medium}  & \multirow{2}{*}{-} \\  & & & \\  \cline{2-4}
   & \multirow{2}{*}{\makecell[l]{ Select the first two photos from the album and create a \\ collage using the first template style in Meitu
   }} &  \multirow{2}{*}{Medium}  & \multirow{2}{*}{-} \\  & & & \\  \hline
   
    \multirow{17}{*}{Long-Horizon }   & \multirow{7}{*}{\makecell[l]{ In the Amap app, find the car rental service. Set the \\ pick-up city to Qingdao, the pick-up and drop-off \\ location to May Fourth Square, and the pick-up and return \\ time from 10:00 to 16:00  tomorrow. Sort the available car \\ models by price in ascending order and select the Comfort \\ type. Stop before filling in the order information. The \\ task ends after confirming all details are correct.
   }} &  \multirow{7}{*}{Hard}  & \multirow{7}{*}{-} \\  & & & \\  & & & \\  & & & \\& & & \\ & & & \\& & & \\ \cline{2-4}
   & \multirow{6}{*}{\makecell[l]{ Go to Tonghuashun to view the A-share stock rankings. Add \\ the following as filter criteria: highlights, industry,  \\and announcements. Filter for stocks with excellent \\ profitability, in the medical equipment industry, and \\ restructured within the past month. After filtering, \\ browse the stock with the  highest trading volume.
   }} &  \multirow{6}{*}{Hard}  & \multirow{6}{*}{-} \\  & & & \\  & & & \\  & & & \\  & & & \\& & & \\ \cline{2-4}
   & \multirow{4}{*}{\makecell[l]{ In Baidu Browser, search for images of "mango pudding" \\ and "matcha pudding" separately. Download one image from \\each search result and open the source of the images. Go \\to "My Pictures" to view the downloaded images.
   }} &  \multirow{4}{*}{Hard}  & \multirow{4}{*}{-} \\  & & & \\  & & & \\  & & & \\    \hline
    \multirow{6}{*}{GUI-Reasoning }  &  Go to Toutiao and clear the search history.
    & -  & Easy\\   \cline{2-4}
   & \multirow{2}{*}{\makecell[l]{ View the details of the first product in the men's \\ clothing tops and jackets category on Pinduoduo.
   }} &  \multirow{2}{*}{-}   & \multirow{2}{*}{Medium} \\  & & & \\  \cline{2-4}
& Play all the news for the watchlist stocks on Tonghuashun.
    & -  & Hard\\   \cline{2-4}
   & \multirow{2}{*}{\makecell[l]{ Go to Jinri Toutiao to change the profile picture, \\ selecting one from the default artistic avatars.
   }} &  \multirow{2}{*}{-}   & \multirow{2}{*}{Hard} \\  & & &   \\\hline\hline
    \end{tabular}
}
    \caption{Task examples and their corresponding difficulty levels}
    \label{table_appendix_tasks}
\end{table*}

\begin{figure*}[!tbh]
\centering
\includegraphics[width=1\textwidth]{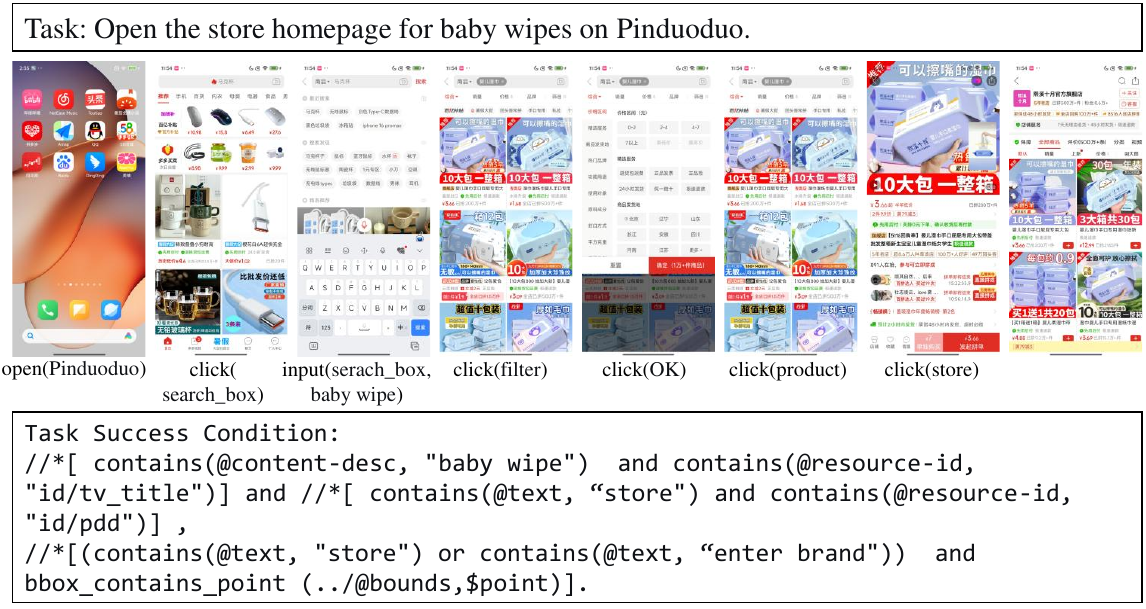} 
\caption{A task example from Base subset and its corresponding success condition.}
\label{figure7_3}
\end{figure*}

\begin{figure*}[!tbh]
\centering
\includegraphics[width=1\textwidth]{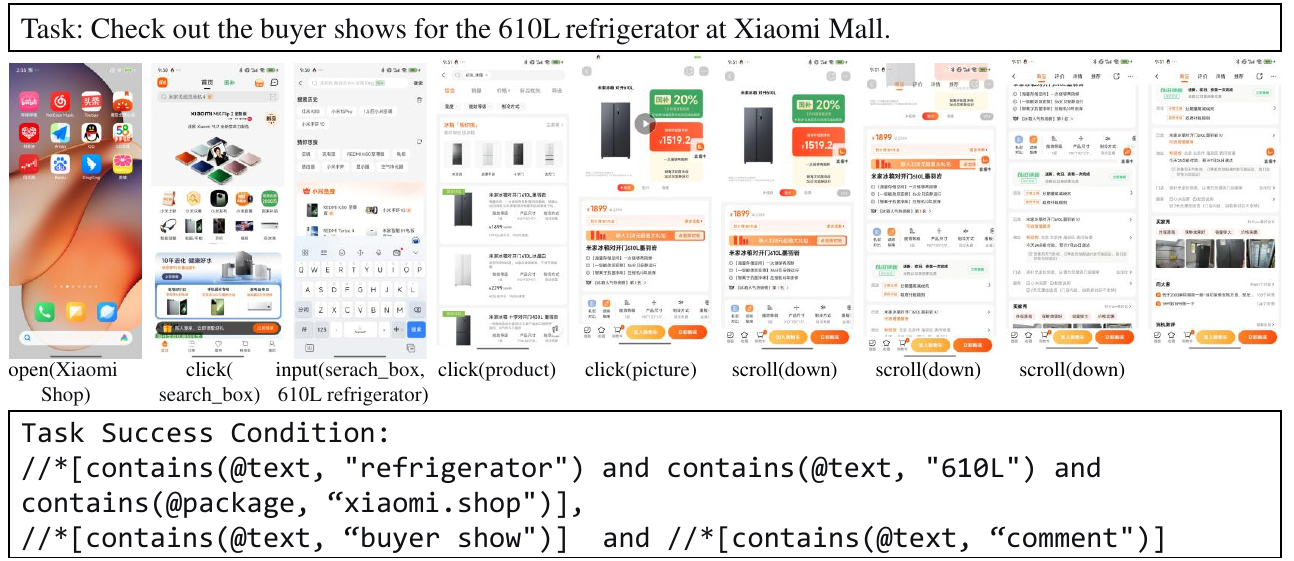} 
\caption{A task example from Long-Tail subset and its corresponding success condition.}
\label{figure7_4}
\end{figure*}

\subsection{Environmental Data} 
The GUI Agent gathers environmental data from mobile devices, which includes screenshots and corresponding XML files. This XML provides a structured hierarchy of the GUI elements, where each node corresponds precisely to a visual element in the screenshot and contains attributes such as text, resource-id, and class.
As shown in the Figure \ref{figure_appendix_6_2}, an element with the text \texttt{My Favorite}, its resource-id is \texttt{tv.danmaku.bili:id/title}, class is \texttt{android.widget.TextView}, content-desc is empty,  package is \texttt{tv.danmaku.bili}, selected is \texttt{false}, clickable is \texttt{true}, bounding box is \texttt{[565,1056][786,1105]}.

When constructing the task success conditions, if we want to represent the "My Favorites" caption element, the importance of each attribute varies. For completeness and soundness, we need to ensure that the attribute does not change every time this element is selected; and that the attributes combined uniquely identify this specific element. Therefore, the text attribute is the most important; the resource-id and package attributes can only filter a subset of elements and are of less important; class, content-desc, and clickable are useless; bounds cannot be used because it cannot be guaranteed that "My Favorites" will be in the same position every time the user enters the profile page.
Therefore, the condition format of this element can be:

$\bullet$ \texttt{@text="My Favorites" and @resource-id = "tv.danmaku.bili:id/title"}

$\bullet$ \texttt{contains(@text, "My Favorites") and contains(@resource-id, "id/title")}

$\bullet$ \texttt{contains(@text, "My Favorites") and @clickable="true" and contains(@package, "bili")  and contains(@resource-id, "title")}

For a task such as "Navigate to My Favorites," there are intuitively two ways to define its success conditions:
(1) Determine whether the My Favorites page appears in the trajectory, and
(2) Determine whether the My Favorites button is clicked on pages such as the Personal Center or System Settings.

For identifying the My Favorites page, the method is to locate elements that are guaranteed to appear on that page. Use multiple elements to uniquely identify the page and avoid similar elements on other pages from triggering false positives. For example, if there is a \texttt{Favorites} element on the page with a resource-id of type \texttt{tab\_title} and selected set to \texttt{true}, and a \texttt{Following} element with a resource-id of type \texttt{tab\_title} and selected set to \texttt{false}, then we can consider this page to be the My Favorites page. The task success condition is:

$\bullet$ \texttt{//*[contains(@text, "Favorites") and @selected="true" and contains(@resource-id, "tab\_title")] and //*[contains(@text, "Following") and @selected="false" and contains(@resource-id, "tab\_title")]}

For detecting a click on the My Favorites button, the approach is to first identify the page preceding the My Favorites page, uniquely distinguishing it using the My Favorites button and other elements on that page. Then, add an action position constraint for the current step to determine whether the action involved interacting with the My Favorites button. Note that accessing the My Favorites page may involve clicking either the corresponding text or the star icon next to it. Therefore, the position constraint is not the TextView bounding box shown in Figure \ref{figure_appendix_6_2}(a), but the ViewGroup bounding box containing both elements, as illustrated in Figure \ref{figure_appendix_6_2}(b). The task success condition is:

$\bullet$ \texttt{//*[contains(@text, "My Favorites") and contains(@resource-id, "id/title") and bbox\_contains\_point (../@bounds, \$point)] and 
//*[contains(@text, "History") and contains(@resource-id, "id/title")]}

\begin{figure}[!tb]
\centering
\includegraphics[width=0.95\columnwidth]{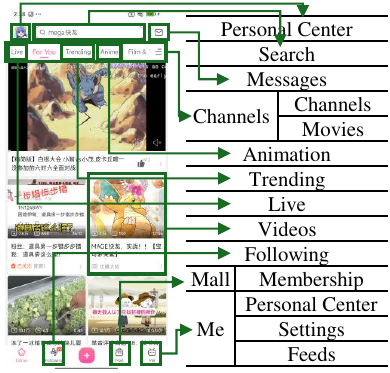} 
\caption{Homepage and 13 key functional points of bilibili. Note that some functions are only accessible via secondary pages.}
\label{figure_appendix_3_2_3}
\end{figure}

\section{MobileBench-OL Benchmark Details}\label{appendix_b}

\subsection{Benchmark Subsets}
Evaluating a GUI agent requires more than completing page navigation tasks on a popular app; it also demands handling long-horizon tasks, exploration, and robustness to noise. We categorize the evaluation into three core dimensions: 1- Base Capabilities (the Base and  Long-Tail subsets), 2- Complex Reasoning (the Long-Horizon and GUI-Reasoning subsets), and 3- Robustness (the Noise-Robust subset). MobileBench-OL leverages this structure to offer a comprehensive evaluation of GUI agents.

Table \ref{table_appendix_tasks} shows several examples for the Base,  Long-Tail, Long-Horizon and GUI-Reasoning subsets.
Figure \ref{figure7_3} and Figure \ref{figure7_4} show examples from the Base subset and Long-Tail subset, respectively.

\subsubsection{Functional Point Coverage}\label{appendix_functional}
In Base subset, when constructing tasks for an app, we expect the benchmark to comprehensively cover all key functional points, not just the core functions. As illustrated in Figure \ref{figure_appendix_3_2_3}, the core function of the Bilibili app is keyword-based video search. However, the app also includes additional functions, such as Channels (movies, videos, trending), Personal Center, and Settings. Ensuring coverage of these secondary functions enables a more comprehensive evaluation of the GUI Agent’s universality and adaptability.

The key functions were identified based on a large number of tasks requested by real users of mobile assistants and are primarily located on the app's first and second-level pages.

Bilibili's functions include: 
\begin{tcolorbox}[    
    colback=yellow!3!white,     
    colframe=yellow!40!black,
breakable, 
                  ]
\begin{verbatim}
|-- Homepage
|   |-- Search
|   |-- Live
|   |-- Animation
|   |-- Movies
|   `-- Sports
|-- Following
|-- Posting Works
|-- Member Shopping
|-- My Profile
|   |-- My Profile
|   |-- Settings
|   |-- Offline Cache
|   |-- History
|   |-- My Favorites
|   |-- Watch Later
|   |-- Scan
|   |-- Comics
|   `-- Member Center
`-- Video Playback Page
    `-- One-Click Three-Fold
\end{verbatim}

\end{tcolorbox}

NetEase Cloud Music's functions include:

\begin{tcolorbox}[
    colback=yellow!3!white,     
    colframe=yellow!40!black,  
                  ]
\begin{verbatim}
|-- Recommendations
|   |-- Search
|   |-- Song Recognition
|   |-- Daily Recommendations
|   |-- Hot Songs Chart
|   |-- Audiobooks
|   |-- Playback
|   |-- Personal Radar
|   |-- Play Cards
|   |-- Podcasts
|   `-- Scan
|-- Personal Roaming
|-- My Profile
|   |-- My Profile
|   |-- Heartbeat Mode
|   |-- Favorites
|   `-- Local
`-- Extended Menu
    |-- Member Center
    |-- Settings
    `-- Timed Shutdown
\end{verbatim}
\end{tcolorbox}

In the Base subset, the annotator first writes at least one query for each function in an app. Then, for a selection of popular and frequently used tasks, they complete all associated queries for that app. This ensures that all of the app's key functionalities are covered.
  
To reduce the number of functionality categories in the dataset, some functions were renamed and merged. However, each consolidated functionality is still guaranteed to have at least one query. For example, "Personal Center" and "My Account" are now collectively referred to as "Personal Center."

\begin{figure*}[!tbh]
\centering
\includegraphics[width=1\textwidth]{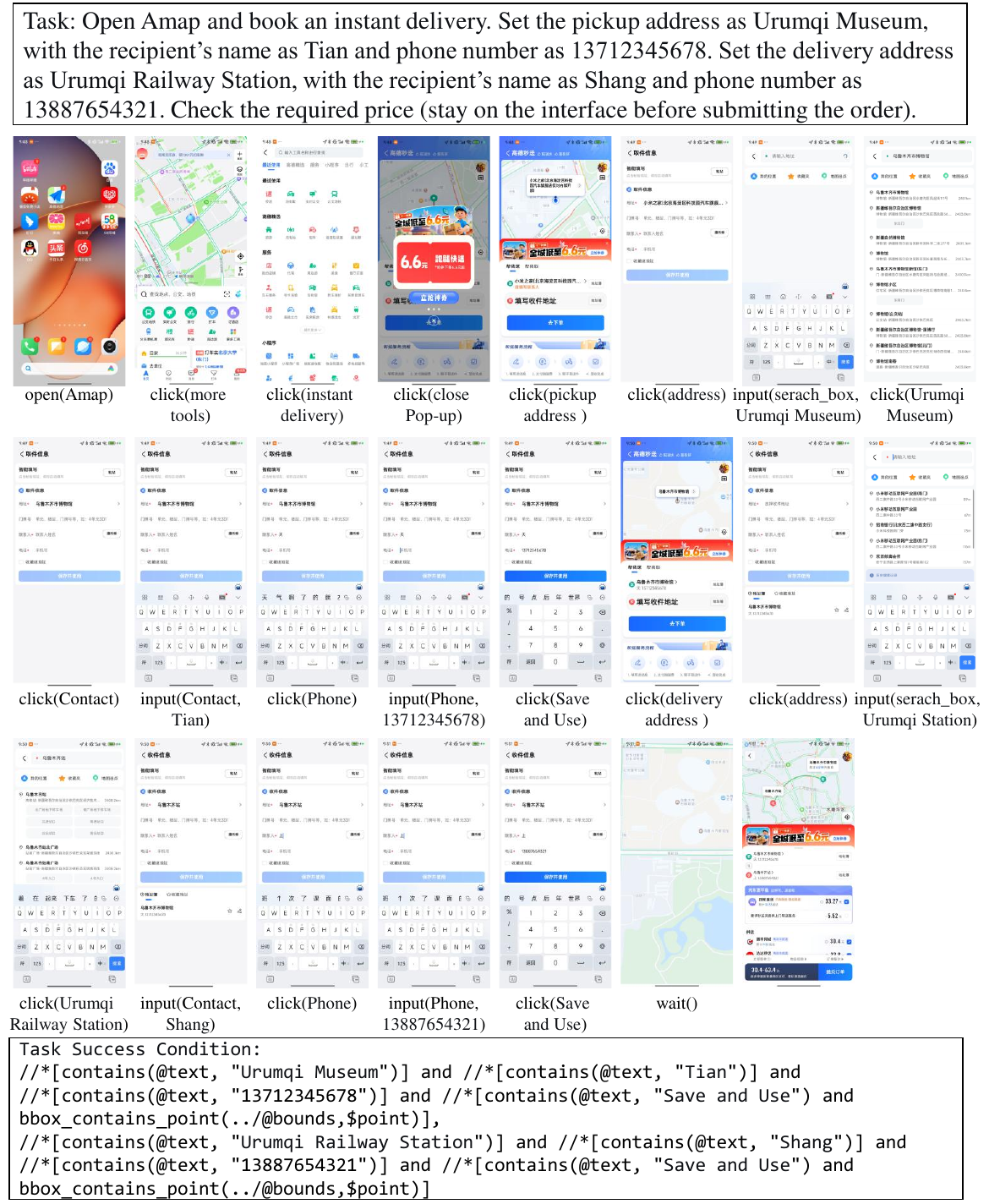} 
\caption{A 22-step long-horizon task and its corresponding success condition.}
\label{figure7_1}
\end{figure*}
\begin{figure*}[!tbh]
\centering
\includegraphics[width=1\textwidth]{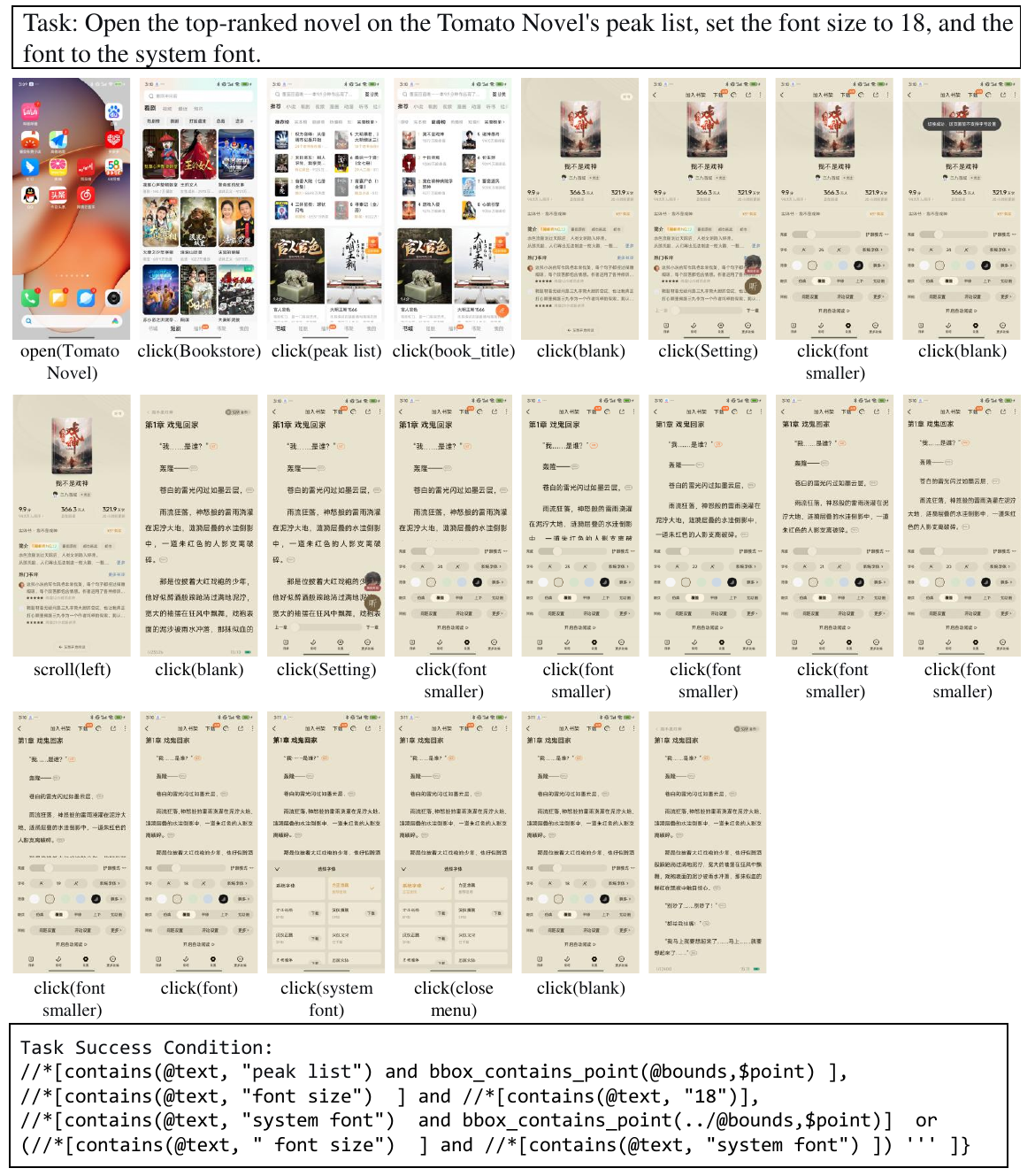} 
\caption{A 21-step long-horizon task and its corresponding success condition.}
\label{figure7_2}
\end{figure*}
\subsubsection{Long-Horizon Subset}

The Long-Horizon subset evaluates the agent's ability to maintain a high-level goal, decompose it into a correct sequence of sub-tasks, and execute them without losing track of the objective. Each task requires at least 20 steps to complete.

We include two examples in Figure \ref{figure7_1} and Figure \ref{figure7_2}.
The task of instant delivery requires filling in multiple fields of information for both the recipient and the sender, testing the ability to manage and sequence multiple subtasks without omission. Adjusting the font size in a novel reader requires repeatedly clicking the reduce button, testing the agent's precise control over components. Other components requiring precise control include date selection, alarm settings, and speed adjustments. In summary, long-horizon tasks require agents to accurately decompose and prioritize subtasks, while keeping track of which subtasks have been completed, which are in progress, and which are still pending.

\subsubsection{GUI-Reasoning Subset}
The GUI-Reasoning subset emulate real-world scenarios that require active exploration and sophisticated reasoning. In these tasks, the goal is not directly achievable through a single, obvious action; instead, the agent must infer how to proceed by understanding the visual context, learning through trial and error, and recovering from dead ends, much like a human user. For example, to "disable messages from strangers," a user wouldn't find this option directly on the main page. They would logically navigate to the "Settings" or "Message" section and browse through sub-menus to locate it.
Similarly, an agent need to explore the interface, testing different paths and recovering from dead ends, which directly tests its exploratory and reasoning abilities.

\begin{figure*}[!t]
\centering
\includegraphics[width=1\textwidth]{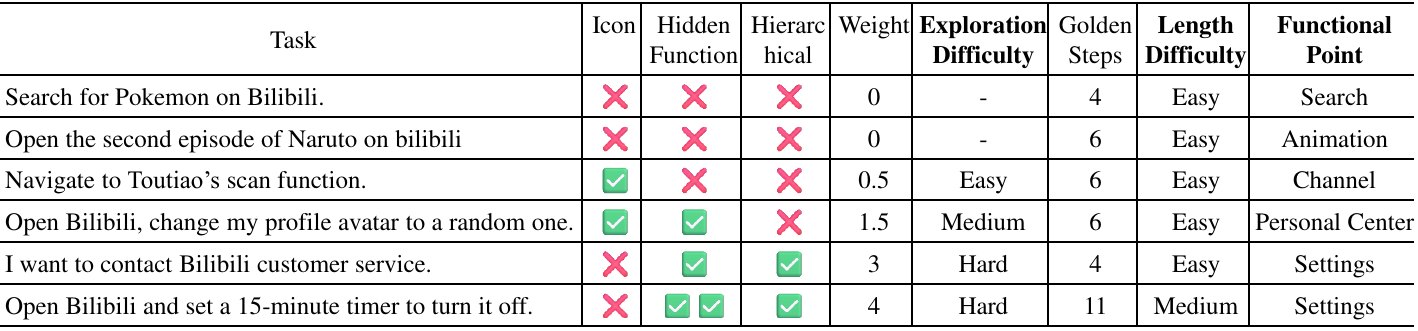} 
\caption{6 MobileBench-OL's example tasks. }
\label{figure3_5}
\end{figure*}

These tasks test advanced exploration capabilities, which we categorize into three key areas: Icon Understanding, Hidden Function Discovery, and Hierarchical Navigation. 
Table \ref{table_2} shows the descriptions and weights of each ability.

$\bullet$ Icon Understanding: Interpreting the meaning of visual elements without explicit text labels.

$\bullet$ Hidden Function Discovery: Locating and accessing functionalities that are not immediately visible on the screen (e.g., finding a "Draft" channel hidden in the submenu, or swiping up to reveal a "Settings" option that is off-screen).

$\bullet$ Hierarchical Navigation: Efficiently traversing deep and complex menu structures to find a specific setting or page. (e.g., finding the "Auto Downloads" option within "Download Settings" in the Settings menu.)

Here, Hidden Function Discovery is page-level, meaning the function is hidden within the current page. Examples include the "More Channels" list displayed by clicking a dropdown button, a pop-up window that appears after clicking "More," or a settings button revealed by swiping up. These functions can be discovered through observation and inference based on the current page.
Hierarchical Navigation is app-level, requiring an understanding of the app's structure. For example, finding the "Timer" function might require navigating to User Center → Settings → General Settings → Timer. Similarly, finding "Stranger Message Notifications" could involve going to User Center → Messages → Options → Notifications. In many cases, the agent might select the wrong element on the User Center page, making the corresponding function difficult to locate.
In this case, Hierarchical Navigation is more difficult; therefore, we assign it a higher difficulty weight.

\begin{table}[!t]
    \centering
\resizebox{1\columnwidth}{!}
{
    \begin{tabular}{c|c|l}
    \hline\hline
       Ability  & \!\!Weight\!\! & Description \\  \hline 
        Icon 	&  \multirow{2}{*}{0.5} 	& Identify and understand icons, \\ 
       Understanding & & graphics, or other visual elements.\!\!\\   \hline 
       \!\!\!\!\!\! Hidden Function \!\!\!\! 	& 	 \multirow{2}{*}{1}	&  Find hidden or advanced features\!\!\\ 
        Discovery&&   on the current page.  \\  \hline 
        Hierarchical  &  \multirow{2}{*}{2} & Understand app structure and   \\  
        Navigation& & hierarchy to find specific functions.\!\!\\ \hline\hline
    \end{tabular}}
    \caption{The descriptions of exploration abilities.
    }
    \label{table_2}
\end{table}

\textbf{Exploration Difficulty Levels:}
The abilities of Icon Understanding, Hidden Function Discovery, and Hierarchical Navigation have assigned weights of 0.5, 1, and 2, respectively.
A task's exploration difficulty is calculated by summing the weights of all abilities required for its completion, as defined by:
$$ \rm Level=\left\{ 
\begin{array}{ll} 
\rm Easy      &  {\rm 0  <  total\;weight \leq 1}\\
\rm Medium    & {\rm 1 <  total\;weight \leq 2}\\
\rm Hard     & {\rm 2 <  total\;weight }
\end{array} \right. $$
Figure \ref{figure3_5} shows several task examples. 
Easy-level task like navigating to scan function (Weight: 0.5), require Scan icon recognition.
Medium-level task, like changing an avatar (Weight:1.5), involve recognizing avatar picture and managing pop-up windows.
Hard-level task, like contacting customer service (Weight:3), has fewer golden steps comparing to change an avator (4<6), but has a higher exploration difficulty because it requires to understand the app structure to locate the customer service area, and scroll up the page to find a hidden feature. Similarly, the task of setting a timer (Weight:4) requires an understanding of the hierarchical structure and finding two hidden functions, Setting and Timer.

\subsubsection{Noise-Roubst Subset}\label{appendix_noise_robust}
The Noise-Roubst subset introduces four distinct types of environmental noise to evaluate agent robustness:

\paragraph{Noise Cause:}
Each noise type is triggered by a specific failure or interruption in the inference process.

$\bullet$ Repeat noise means the agent incorrectly assumes a previous action has failed. This is typically caused by insufficient waiting time after action execution or by fetching the device status too quickly, leading the agent to generate and execute the same action repeatedly.

$\bullet$ Unexecuted noise means the agent’s previous action fails to trigger the expected UI response (e.g., a button click that has no effect on the interface).

$\bullet$ Delay noise arises from extended interruptions that prevent state progression. This includes Page Loading Delays, temporary Network Disconnection, or Ad-delay pages that hold the interface in a loading state.

$\bullet$ Pop-up noise is triggered by unexpected overlay pages that interrupt the main workflow. These include Login Pages requiring authentication, Captcha Pages with unsolvable challenges, Ad Pop-ups obscuring the UI, and Cookie Consent Pop-ups that must be dismissed.

\paragraph{Handling Method:}
The required agent response and the potential negative outcomes of an incorrect response are closely linked for each noise type.

$\bullet$ For Repeat noise, the agent must return to the previous page. If mishandled, the page will have already advanced after the second execution, forcing the agent to restart from a new, unintended state. This may be misinterpreted as the initial action causing two page transitions.

$\bullet$ For Unexecuted noise, the agent must re-execute the intended action. Failure to do so results in an unchanged state, requiring the agent to re-observe and retry. This can be mistaken for the action itself being invalid.

$\bullet$ For Delay noise, the correct response is to wait. If the agent waits correctly, the true result page loads normally. However, if the agent generates any other action during the delay, that action will be performed on the real result state (which is hidden), often leading to errors or navigation to an incorrect page.

$\bullet$ For Pop-up noise, the agent must close the pop-up. A correct click on the close button will reveal the true underlying page. If the agent performs any other action, the pop-up will persist in the agent's view, and the action will have no effect on the real device state, creating a deadlock.

\paragraph{Noise Injection Method:}
The noise injection method and the page state presented to the agent are as follows.

For Repeat noise, the injection method is to execute the same agent action twice consecutively. The agent is then shown the page that results from this double execution.

For Unexecuted noise, the injection method is to block the agent’s action from being passed to the device for execution. Consequently, the agent continues to see the unchanged previous page.

For Delay and Popup noises, a special simulation approach is used. Because it is impossible to reliably source random real-world waiting or pop-up pages, pre-prepared template pages are used. The noise injection method involves copying and presenting these pre-prepared pages to the agent. Meanwhile, the device continues to process the action in the background and reaches the real result state. Therefore, the agent is shown a simulated Delay (waiting) page or a simulated Popup page, while the device displays the true outcome of the action.

\subsection{Constuction Process}
We manually design tasks from scratch for each app, rather than modifying existing datasets, to reduce the risk of large language model (LLM) data contamination. During this process, we focus on the agent's task execution and exploration abilities, while ensuring diversity in tasks in terms of difficulty levels and function coverage.

Three human annotators constructed tasks based on the apps from scratch. 
When constructing tasks, we begin by analyzing the app’s hierarchical structure, page layout, and distribution of function points. For the high-frequency "Base" subset, we annotate 20–30 tasks per app, ensuring comprehensive coverage of all function points. For the "Long-Tail" subset (with lower frequency), we annotate 5 tasks per app. Since their core functions differ significantly from those of the Base apps, we mainly focus on their core functions to enhance the benchmark’s comprehensiveness.  For the 'Long-Horizon' subset, we ensure that the tasks within it require at least 20 steps to be completed.  For the 'GUI-Reasoning' subset, we ensure that the tasks within it require at least one exploration task to be completed. Noise-Robust use the same test set as the Base subset.

\begin{figure*}[!tbh]
\centering
\includegraphics[width=0.8\textwidth]{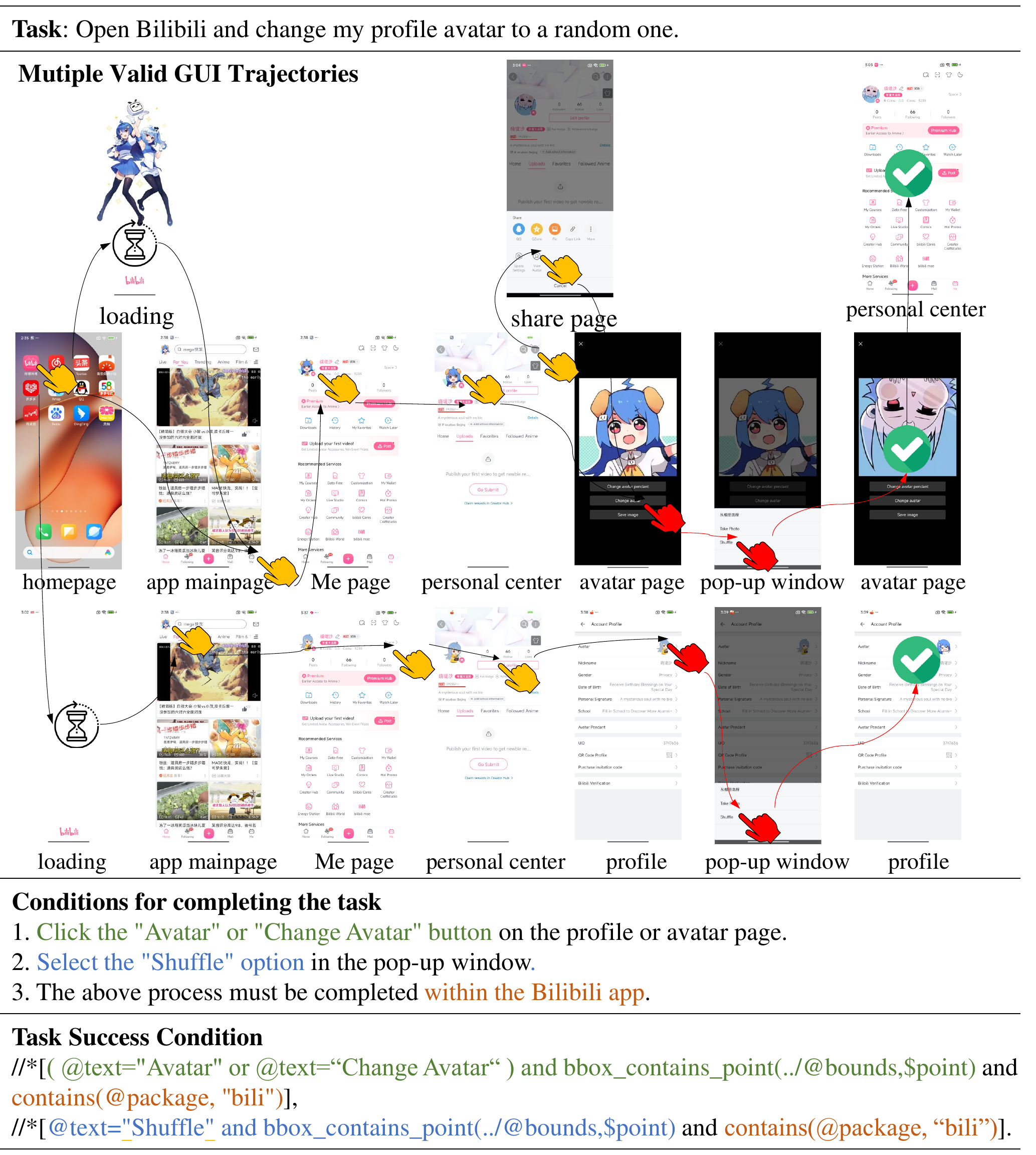} 
\caption{An example task and its corresponding success condition. Red arrows indicate the definitive indicators for completing the task.}
\label{figure1_2}
\end{figure*}

\subsection{Dataset statistics}\label{appendix_dataset_statistics}
Table \ref{table_4} shows the dataset statistics for MobileBench-OL.

\paragraph{Task Distribution:} 
MobileBench-OL is divided into five subsets.
The Base subset contains 310 tasks covering 12 mainstream apps, with 20–30 tasks per app.
The Long-Tail subset contains 340 tasks covering 68 less popular apps, with 5 tasks per app.
The Long-Horizon subset contains 60 tasks, with 5 tasks for each of the 12 top apps.
The GUI-Reasoning subset also contains 60 tasks, with 5 tasks per app.
The Noise-Robust subset follows the same task distribution as the Base subset but explicitly introduces noise only during inference. Due to significant fluctuations in golden steps caused by noise, the average steps are not recalculated after noise is introduced.

\paragraph{Difficulty Distribution:} 
The difficulty of the Base, Long-Tail, Long-Horizon, and Noise-Robust subsets is categorized based on their golden steps. In Long-Horizon, all tasks require more than 20 steps to complete and are classified as Hard. The GUI-Reasoning subset rates task difficulty according to the required exploration ability and contains contains more Medium and Hard tasks.

\paragraph{Functional Point Distribution:}
The Base subset covers all 28 core functional points of the top apps. The Long-Tail and GUI-Reasoning subsets cover fewer functional points, focusing on core functions of the long-tail apps and exploratory tasks, respectively.

\begin{table*}[!t]
    \centering
\resizebox{1\textwidth}{!}{
    \begin{tabular}{l|l}
    \hline\hline
    Tasks & Task Success Conditions \\  \hline
    Open Bilibili and change my     & //*[(@text="Avatar" or @text=“Change Avatar“ ) and bbox\_contains\_point  \\
    profile avatar to a random one.\!\! & (../@bounds, \$point) and contains(@package, "bili")] and //*[@text="Shuffle" \\
    &  and bbox\_contains\_point(../@bounds,\$point) and contains(@package, “bili”)].  \\ \hline
    I want to contact Bilibili &
    //*[(@text="Customer Service" or @text=“Help Center“) \\
    customer service. & and bbox\_contains\_point(../@bounds, \$point) and contains(@package, "bili")]. \\ \hline
    Find the subway route from  & //*[contains(@text, "Public Transportation") and @selected="true" and contains\\ 
    Beijing South Railway Station  & (@package, "map")] and //*[contains(@text, "Beijing South Railway Station")  and \\
    to Beijing Fengtai Station \!\!  &   contains(@resource-id, "route\_edit\_summary\_start")] and //*[contains(@text,  \\
    on Amap. & "Beijing Fengtai Station") and  contains(@resource-id,"route\_edit\_summary\_end")] \\\hline\hline
    \end{tabular}}
    \caption{Example tasks and their corresponding success
conditions.}
    \label{table_appendix_3}
\end{table*}

\begin{figure*}[!tbh]
\centering
\includegraphics[width=1\textwidth]{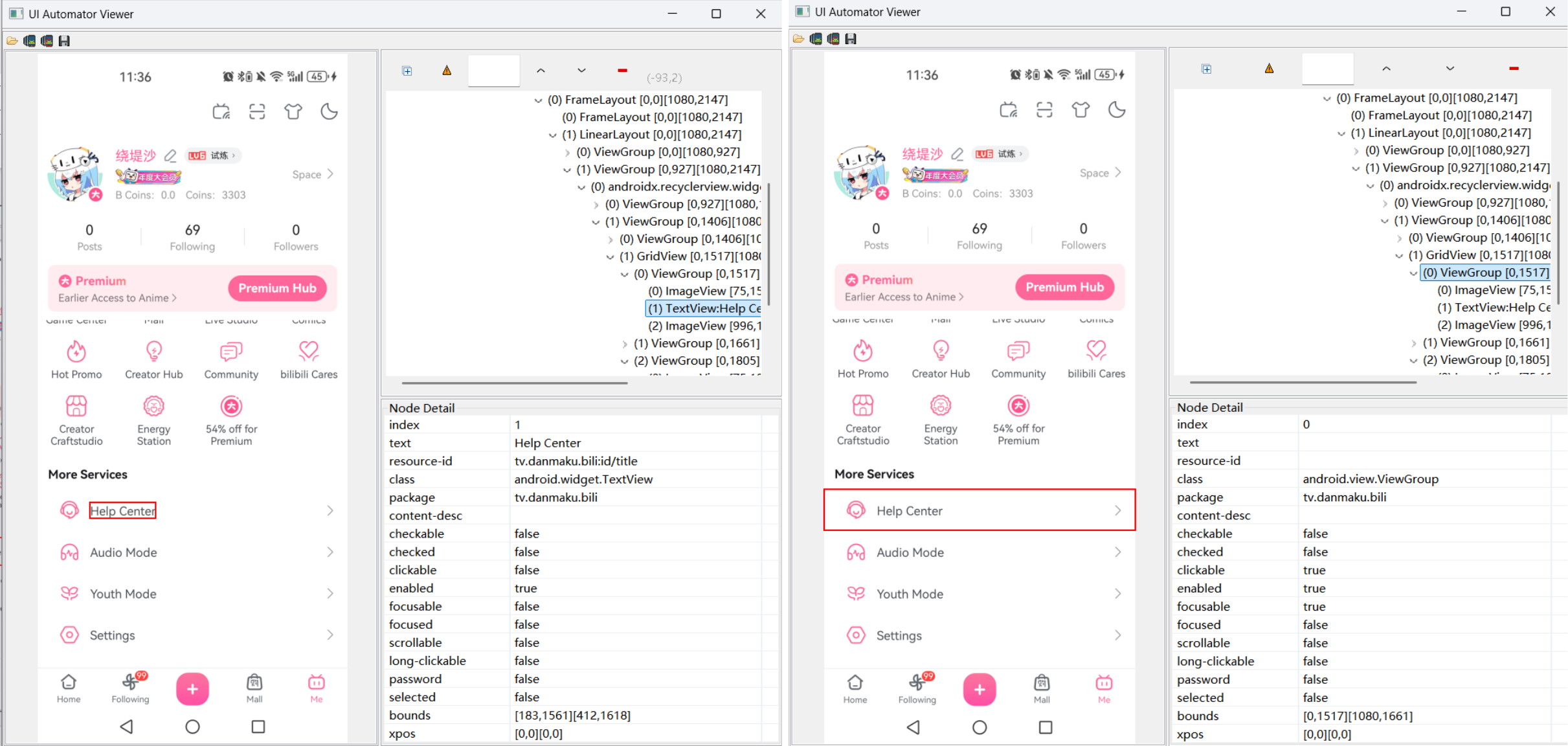} 
\caption{The bounding box of the 'Help Center' and the parent node of 'Help Center'.}
\label{figure_appendix_6_4}
\end{figure*}

\section{Data Analysis}\label{appendix_condition}
\subsection{GUI Trajectory}
In real-world environments, a task may correspond to multiple valid GUI trajectories, as shown in Figure \ref{figure1_2}. We have human annotators label a golden trajectory for each task. Here, annotators remove loading screens, ad pop-ups, permission request pages, etc., and focus only on completing the task itself. The middle row of trajectories in Figure \ref{figure1_2} can be regarded as the golden trajectory for this task. It is worth noting that the golden trajectory is not necessarily always the shortest valid path. For example, to find Customer Service in Figure \ref{figure1_1}, the shortest path requires only 4 steps, but people usually do not realize that the Help Center also navigates to Customer Service, and may take 7 steps to find the Customer Service function through the Settings.

\begin{table*}[!t]
    \centering
\resizebox{1\textwidth}{!}{
    \begin{tabular}{l|l|l}
    \hline\hline
    Original Task &
 Category
& Reset Task \\  \hline
   \multirow{2}{*}{\makecell[l]{Turn on Pinduoduo's\\senior mode.}}  & \multirow{2}{*}{\makecell[c]{Task-level\\ Reset}}
    & Turn off Pinduoduo's senior mode. Follow these steps: 1. Open \\  & & Pinduoduo and click the "Personal Center" button in the lower right 
    \\ & & corner of the homepage to go to the personal center page. 2. If you see a 
    \\ & & "Close Large Text" button in the upper right corner, it means Senior
    \\ & &  Mode is enabled, click it. If the button is not there, Senior Mode is 
    \\ & & already off, and no further action is needed. 3. On the Senior Mode page, 
    \\ & & click "Close Senior Mode" to complete the process. Follow the above
    \\ & &  steps to turn off Pinduoduo's senior mode.\\ \hline
   \multirow{3}{*}{\makecell[l]{Turn on Bilibili \\  private message  \\intelligent filtering.}} & \multirow{2}{*}{\makecell[c]{Task-level\\ Reset}}
    & I need to turn off the private message intelligent filtering on Bilibili. 
    \\  & & Follow these steps: 1. Open Bilibili and tap the "My" button at the bottom
    \\ & & of the  home page. 2. Once on the "My" page, if the background is black,
    \\ & & tap the sun  icon in the top right to switch to light mode. If the background
    \\ & &  is already white, proceed to step 3. 3. On the "My" page, swipe up until
    \\ & & you see the  "Settings" button at the bottom of the More Services section. 
    \\ & & 4. Tap the "Settings" button to enter the Settings page. 5. On the Settings
    \\ & & page, tap  "Message Settings" to go to the Message Settings page. 6. In the
    \\ & & Message  Settings page, if the toggle next to "Private Message Intelligent
    \\ & & Filtering" is  turned on, tap it to turn it off. If the toggle is already off, the
    \\ & & intelligent  filtering feature is disabled, and the task is complete.
    \\ \hline
   \multirow{3}{*}{\makecell[l]{Search for today's\\stock price of \\Wuliangye on Toutiao.}} 
 & \multirow{2}{*}{\makecell[c]{App-level\\Reset}}
    & Clear Toutiao search history with the following steps: 1. Open Toutiao
    \\   & &  and tap the search box at the top of the home page. 2. In the search page,
    \\  & &  there is a trash can icon on the right side of the search history section—
    \\\cline{1-1}
    \multirow{2}{*}{\makecell[l]{Search for today's\\gold price on  Toutiao.}}  & & this represents clearing search history. Tap this trash can button. 3. In the 
    \\ 
     & & expanded options, tap the "Delete All" button. 4. In the pop-up window 
    \\  & & asking "Are you sure you want to clear the search history?", tap the 
    \\  & & "Confirm" button. Follow the steps above to clear the search history.\\\hline
    \multirow{4}{*}{\makecell[l]{Check out the first news\\ article in the Toutiao \\Picture Channel and \\give it a like.}}  & \multirow{2}{*}{\makecell[c]{No reset\\needed }}&  -\\
     & &  \\
     & &  \\
     & &  \\  \hline
    \multirow{3}{*}{\makecell[l]{Open Bilibili messages\\ and mark  all as read \\with one click.}} & \multirow{1}{*}{Infeasible}& - \\
     & &  \\
     & &  \\
\hline\hline
    \end{tabular}}
    \caption{Example reset tasks from different Reset Categories.}
    \label{table_appendix_4}
\end{table*}
\subsection{Task Success Conditions}
To determine whether a GUI trajectory successfully completes a task in MobileBench-OL, we define task success conditions. These conditions may consist of multiple sub-conditions involving element matching (verifying the presence of specific GUI elements) and action matching (checking for critical actions). Each condition is annotated using a rule-based approach to ensure robustness.

Since a single task may have multiple valid GUI trajectories, we construct success conditions using only those elements and actions that appear exclusively in successful trajectories. This ensures: (1) Completeness: Every successful trajectory satisfies the entire condition. (2) Soundness: No unsuccessful trajectory fully meets the condition. This approach guarantees that the evaluation is both precise (only correct trajectories pass) and flexible (matching diverse but valid paths).

For example, in Figure \ref{figure1_2}, the success condition of task "Change to a random Bilibili avatar" is clicking the Avatar button and then selecting Shuffle, regardless of the navigation path. This design provides precise evaluation while tolerating diverse valid paths.

To ensure scalability and stability, task success conditions are uniformly defined using XPath-like rules. Each GUI element is characterized by attributes such as text, resource-id, class, package, and checked. These attributes are used to describe the required elements and actions in the rules. Finally, task success conditions are converted into normalized format.

As shown in Figure \ref{figure1_2}, the first sub-condition uses "contains(@package, "bili")" to confirm the current page belongs to bilibili, text="Avatar" ensures the presence of the "Avatar" element, and bbox\_contains\_point verifies that the interaction occurs within the bounding box of the "Avatar" button. Similarly, the second sub-condition checks for an element with the text "Shuffle" and the package "bili", and the current action occurs within the bounding box of this element.

Table \ref{table_appendix_3} shows more examples. For the task of accessing customer service, clicking on the Help Center or Customer Service can accomplish it. It is worth noting that '../@bounds' refers to the parent node of this element. The three attribute constraints of this element in the Task Success Conditions are described in natural language as follows: the text of this element contains 'Help Center' or 'Customer Service', the package includes 'bili', and the interaction point for the next action on this GUI page falls within the bounding box of this element's parent node. As shown in Figure \ref{figure_appendix_6_4}, the bounding box of the element with the text 'Help Center' covers only the text box, while the actual boundary for accessing this function should also include the headphone icon on the left and the blank area on the right. Therefore, we need the interaction point to fall within the parent node of 'Help Center'.

\begin{figure*}[!t]  

\begin{tcolorbox}[   
    colback=green!3!white,     
    colframe=green!40!black,
    width=\textwidth,  
    center,
    before={\par\noindent},  
    after={\par}  ]
You are an agent who can operate an Android phone on behalf of a user. Based on user's goal/request, you may
- Complete some tasks described in the request by performing actions on the phone using visual understanding.

At each step, you will be given the history path, current screenshot (before action) and the task goal.
You must analyze the screen and output your action decision:

1. A brief reasoning in Chinese: Why and where to take the next action.

2. A structured action command in format below.

Supported Actions:

- Click/tap a position on screen: `\{\{"click(start\_point=(x1,y1))"\}\}`

- Scroll the screen: `\{\{"scroll(start\_box=(x1,y1), end\_box=(x2,y2))"\}\}`

- Type text into an input field when searching: `\{\{"type(content=...)\}\}`

- Press home button: `\{\{press\_home()\}\}`

- Press back button: `\{\{press\_back()\}\}`

- Wait for UI update: `\{\{wait()\}\}`

- The task is finished: `\{\{finished(content=)\}\}`

You must only use the above 7 actions. 

Use coordinates based on your visual understanding of the screenshot.

The current user goal/request is: \{goal\}

the size of the screenshot is 1080 * 2400

Here is a history of what you have done so far:

\{history\}

The current screenshot and the same screenshot with bounding boxes and labels added are also given to you.

Here is a list of detailed information for some of the UI elements (notice that some elements in this list may not be visible in the current screen and so you can not interact with it, can try to scroll the screen to reveal it first), the numeric indexes are consistent with the ones in the labeled screenshot:
\{ui\_elements\}

Here are some useful guidelines you need to follow:

General:

...

Action Related:

...

Text Related Operations:

...

Now output your decision:
Thought: Please infer your intention and location in Chinese.
Action: (structured action selected from above 7 actions)...
\end{tcolorbox}
\caption{The prompt of GPT-4o.}
\label{Prompt_1}
\end{figure*}

\begin{figure*}[!t]  

\begin{tcolorbox}[   
    colback=green!3!white,     
    colframe=green!40!black,
    width=\textwidth,  
    center,
    before={\par\noindent},  
    after={\par}  ]
You are an agent who can operate an Android phone on behalf of a user. Based on user's goal/request, you may
- Answer back if the request/goal is a question (or a chat message), like user asks "What is my schedule for today?".
- Complete some tasks described in the requests/goals by performing actions (step by step) on the phone.

When given a user request, you will try to complete it step by step. At each step, you will be given the current screenshot (including the original screenshot and the same screenshot with bounding boxes and numeric indexes added to some UI elements) and a history of what you have done (in text). Based on these pieces of information and the goal, you must choose to perform one of the action in the following list (action description followed by the JSON format) by outputing the action in the correct JSON format.

- If you think the task has been completed, finish the task by using the status action with complete as goal\_status: `\{\{"action\_type": "status", "goal\_status": "complete"\}\}`

- If you think the task is not feasible (including cases like you don't have enough information or can not perform some necessary actions), finish by using the `status` action with infeasible as goal\_status: `\{\{"action\_type": "status", "goal\_status": "infeasible"\}\}`

- Answer user's question: `\{\{"action\_type": "answer", "text": "<answer\_text>"\}\}`

- Click/tap on an element on the screen. We have added marks (bounding boxes with numeric indexes on their TOP LEFT corner) to most of the UI elements in the screenshot, use the numeric index to indicate which element you want to click: `\{\{"action\_type": "click", "index": <target\_index>\}\}`.

- Long press on an element on the screen, similar with the click action above, use the numeric label on the bounding box to indicate which element you want to long press: `\{\{"action\_type": "long\_press", "index": <target\_index>\}\}`.

- Type text into a text field (this action contains clicking the text field, typing in the text and pressing the enter, so no need to click on the target field to start), use the numeric label on the bounding box to indicate the target text field: `\{\{"action\_type": "input\_text", "text": <text\_input>, "index": <target\_index>\}\}`

- Press the Enter key: `\{\{"action\_type": "keyboard\_enter"\}\}`
- Navigate back: `\{\{"action\_type": "navigate\_back"\}\}`

- Scroll the screen or a scrollable UI element in one of the four directions, use the same numeric index as above if you want to scroll a specific UI element, leave it empty when scroll the whole screen: `\{\{"action\_type": "scroll", "direction": <up, down, left, right>, "index": <optional\_target\_index>\}\}`

- Wait for the screen to update: `\{\{"action\_type": "wait"\}\}`

The current user goal/request is: \{goal\}
Here is a history of what you have done so far:
\{history\}
The current screenshot and the same screenshot with bounding boxes and labels added are also given to you.

Here is a list of detailed information for some of the UI elements (notice that some elements in this list may not be visible in the current screen and so you can not interact with it, can try to scroll the screen to reveal it first), the numeric indexes are consistent with the ones in the labeled screenshot:
\{ui\_elements\}

Here are some useful guidelines you need to follow:
...
\{additional\_guidelines\}

Now output an action from the above list in the correct JSON format, following the reason why you do that. Your answer should look like:

Reason: ...

Action: \{\{"action\_type":...\}\}

Your Answer:
\end{tcolorbox}
\caption{The prompt of M3A.}
\label{Prompt_2}
\end{figure*}

\begin{figure*}[!t]  

\begin{tcolorbox}[   
    colback=green!3!white,     
    colframe=green!40!black,
    width=\textwidth,  
    center,
    before={\par\noindent},  
    after={\par}  ]
You are an agent who can operate an Android phone on behalf of a user. Based on user's goal/request, you may

- Answer back if the request/goal is a question (or a chat message), like user asks "What is my schedule for today?".

- Complete some tasks described in the requests/goals by performing actions (step by step) on the phone.

When given a user request, you will try to complete it step by step. At each step, a list of descriptions for most UI elements on the current screen will be given to you (each element can be specified by an index), together with a history of what you have done in previous steps. Based on these pieces of information and the goal, you must choose to perform one of the action in the following list (action description followed by the JSON format) by outputing the action in the correct JSON format.

- If you think the task has been completed, finish the task by using the status action with complete as goal\_status: `\{\{"action\_type": "status", "goal\_status": "complete"\}\}`

- If you think the task is not feasible (including cases like you don't have enough information or can not perform some necessary actions), finish by using the `status` action with infeasible as goal\_status: `\{\{"action\_type": "status", "goal\_status": "infeasible"\}\}`

- Answer user's question: `\{\{"action\_type": "answer", "text": "<answer\_text>"\}\}`

- Click/tap on a UI element (specified by its index) on the screen: `\{\{"action\_type": "click", "index": <target\_index>\}\}`.

- Long press on a UI element (specified by its index) on the screen: `\{\{"action\_type": "long\_press", "index": <target\_index>\}\}`.

- Type text into an editable text field (specified by its index), this action contains clicking the text field, typing in the text and pressing the enter, so no need to click on the target field to start: `\{\{"action\_type": "input\_text", "text": <text\_input>, "index": <target\_index>\}\}`

- Press the Enter key: `\{\{"action\_type": "keyboard\_enter"\}\}`

- Navigate back: `\{\{"action\_type": "navigate\_back"\}\}`

- Scroll the screen or a scrollable UI element in one of the four directions, use the same numeric index as above if you want to scroll a specific UI element, leave it empty when scroll the whole screen: `\{\{"action\_type": "scroll", "direction": <up, down, left, right>, "index": <optional\_target\_index>\}\}`

- Wait for the screen to update: `\{\{"action\_type": "wait"\}\}`

The current user goal/request is: \{goal\}

Here is a history of what you have done so far:
\{history\}

Here is a list of descriptions for some UI elements on the current screen:
\{ui\_elements\_description\}
Here are some useful guidelines you need to follow:
General
...
\{additional\_guidelines\}

Now output an action from the above list in the correct JSON format, following the reason why you do that. Your answer should look like:
Reason: ...
Action: \{\{"action\_type":...\}\}

Your Answer:
\end{tcolorbox}
\caption{The prompt of T3A.}
\label{Prompt_3}
\end{figure*}

\section{Reset Mechanism}\label{appendix_reset}
Table \ref{table_appendix_4} illustrates several reset tasks from different reset categories. Given an original task such as "Turn on Bilibili private message intelligent filtering," the reset task is a step-by-step inversion designed to turn off "private message intelligent filtering." At each step, we prompt the agent by specifying which element it should interact with, along with the element’s possible location, shape, and the current GUI's likely state. In this way, we guide the agent to reset the device state as completely as possible, regardless of the original task's complexity.

For Task-level resets, each original task is assigned a separate reset task. For App-level resets, multiple original tasks share a single reset task. For example, an original task like "Search for today's stock price of Wuliangye on Toutiao" has the reset task "Clear Toutiao search history." Thus, various search tasks on Toutiao can all use this single reset task, which only needs to be performed once per benchmark test epoch

For tasks that require no reset, such as "Check out the first news article in the Toutiao Picture Channel and give it a like," the process is purely observational. These tasks leave no lasting environmental changes or shortcuts, and the navigation process in one trial does not interfere with the trajectory generation of the next.

For infeasible tasks, such as "Open Bilibili messages and mark all as read with one click," once all unread messages on the device are marked as read, we can no longer accurately identify and revert them to an unread state. Therefore, for such tasks, we consider two success scenarios and define the corresponding success conditions for each: successfully clicking the one-click read button within the trajectory, or arriving at the message page and finding no unread messages.

\begin{figure*}[!t]  

\begin{tcolorbox}[   
    colback=green!3!white,     
    colframe=green!40!black,
    width=\textwidth,  
    center,
    before={\par\noindent},  
    after={\par}  ]
\#\#\# Background \#\#\#

This image is a phone screenshot. Its width is \{width\} pixels and its height is \{height\} pixels. The user's instruction is: \{instruction\}.

\#\#\# Screenshot information \#\#\#

In order to help you better perceive the content in this screenshot, we extract some information on the current screenshot through system files. This information consists of two parts: coordinates; content. The format of the coordinates is [x, y], x is the pixel from left to right and y is the pixel from top to bottom; the content is a text or an icon description respectively. The information is as follow:
\{clickable\_infos\}

Please note that this information is not necessarily accurate. You need to combine the screenshot to understand.

\#\#\# Keyboard status \#\#\#

We extract the keyboard status of the current screenshot and it is whether the keyboard of the current screenshot is activated.
The keyboard status is as follow:
\{keyboard\_status\}

\#\#\# Hint \#\#\#

There are hints to help you complete the user's instructions. The hints are as follow:

The required app is already open. Please do not use the "Open app" or "Home" actions.

\#\#\# History operations \#\#\#

Before reaching this page, some operations have been completed. You need to refer to the completed operations to decide the next operation. These operations are as follow:
\{action\_history\}

\#\#\# Progress \#\#\#

After completing the history operations, you have the following thoughts about the progress of user's instruction completion:
Completed contents:
\{completed\_content\}

\#\#\# Memory \#\#\#

During the operations, you record the following contents on the screenshot for use in subsequent operations:
Memory:
\{memory\}

\#\#\# Response requirements \#\#\#

Now you need to combine all of the above to perform just one action on the current page. You must choose one of the six actions below:

Open app (app name): If the current page is desktop, you can use this action to open the app named "app name" on the desktop.

Tap (x, y): Tap the position (x, y) in current page.

Swipe (x1, y1), (x2, y2): Swipe from position (x1, y1) to position (x2, y2).

\{type\_action\}

Back: Return to the previous page.

Home: Return to home page.

Stop: If you think all the requirements of user's instruction have been completed and no further operation is required, you can choose this action to terminate the operation process.

\#\#\# Output format \#\#\#

Your output consists of the following three parts:

\#\#\# Thought \#\#\#

Think about the requirements that have been completed in previous operations and the requirements that need to be completed in the next one operation.

\#\#\# Action \#\#\#

You can only choose one from the six actions above. Make sure that the coordinates or text in the "()".

\#\#\# Operation \#\#\#

Please generate a brief natural language description for the operation in Action based on your Thought.
\end{tcolorbox}
\caption{The prompt of Mobile-Agent-V2.}
\label{Prompt_4}
\end{figure*}

\begin{figure*}[!t]  

\begin{tcolorbox}[   
    colback=green!3!white,     
    colframe=green!40!black,
    width=\textwidth,  
    center,
    before={\par\noindent},  
    after={\par}  ]
You are a GUI agent. You are given a task and your action history, with screenshots. You need to perform the next action to complete the task. 

\#\# Output Format

Thought: ...

Action: ...

\#\# Action Space

click(start\_box='<|box\_start|>(x1,y1)<|box\_end|>')

type(content='')

scroll(direction='down or up or right or left')

press\_back()

press\_home()

wait()

finished() \# Submit the task regardless of whether it succeeds or fails.

\#\# Note

- Use English in Thought part.

- Summarize your next action (with its target element) in one sentence in Thought part.

\#\# User Instruction
\end{tcolorbox}
\caption{The prompt of UI-TARS.}
\label{Prompt_5}
\end{figure*}

\begin{figure*}[!t]  

\begin{tcolorbox}[   
    colback=green!3!white,     
    colframe=green!40!black,
    width=\textwidth,  
    center,
    before={\par\noindent},  
    after={\par}  ]
You are a GUI agent. You are given a task and your action history, with screenshots. You need to perform the next action to complete the task.

\#\# Output Format

```

Thought: ...

Action: ...

```

\#\# Action Space

click(point='<point>x1 y1</point>')

type(content='') \#If you want to submit your input, use "\textbackslash n" at the end of `content`.

scroll(point='<point>x1 y1</point>', direction='down or up or right or left')

press\_home()

press\_back()

wait()

finished(content='xxx') \# Use escape characters ', ", and \textbackslash n in content part to ensure we can parse the content in normal python string format.

\#\# Note

- Use Chinese in `Thought` part.

- Write a small plan and finally summarize your next action (with its target element) in one sentence in `Thought` part.

\#\# User Instruction
\end{tcolorbox}
\caption{The prompt of Qwen2-VL, Qwen2.5-VL and UI-TARS-1.5.}
\label{Prompt_6}
\end{figure*}

\section{Prompt}\label{appendix_g}

The prompt for GPT-4o is shown in Figure \ref{Prompt_1}. The prompts used in M3A and T3A are provided in Figures \ref{Prompt_2} and \ref{Prompt_3}, respectively. The prompt used in Mobile-Agent-V2 can be seen in Figure \ref{Prompt_4}. The prompt for UI-TARS is in Figure \ref{Prompt_5}. The prompts for Qwen2-VL, Qwen2.5-VL and UI-TARS-1.5 are shown in Figure \ref{Prompt_6}.

\section{More Related Work} \label{appendix_related}

\paragraph{GUI Agents.}
The field of GUI agents has made significant progress recently, driven by the rapid development of Multimodal Large Language Models (MLLMs) \cite{openai2023gpt4,liu2023visual} and large-scale GUI datasets \cite{zhan2023you,wu2024mobilevlm,deng2023mind2web}. 
Some current research \cite{wang2024mobile, lu2024omniparser} leverages powerful closed-source models like GPT-4o \cite{yan2023gpt4vwonderlandlargemultimodal} and Gemini \cite{comanici2025gemini}, using prompt engineering to build GUI agents.
Other work \cite{bai2024digirl, wang2024distrl} employs supervised fine-tuning (SFT) and reinforcement learning from human feedback (RLHF) on open-source base models trained on GUI datasets. 
These studies have benefited from the support of growing large-scale GUI datasets. 
For example, AITW \cite{rawles2023android} and Android Control \cite{li2024effects} support GUI navigation tasks, while AMEX \cite{chai2024amex} and Ferret-UI \cite{you2024ferret} enable deeper understanding of mobile UI hierarchies and elements. 
To enhance GUI agents' ability to handle complex tasks, recent studies \cite{wu2025backtrackagent,wang2025mobileagenteselfevolvingmobileassistant,liu2025infiguiagentmultimodalgeneralistgui} have introduced advanced features such as task planning, memory storage, execution reflection and error correction. Rapidly advancing GUI agents also require comprehensive benchmarks to evaluate their performance across various tasks and environments.

\paragraph{GUI Benchmarks.} With the rapid advancement of GUI Agents, traditional static benchmarks have become inadequate for comprehensively evaluate their full abilities.
Recently, dynamic benchmark environments have been proposed \cite{kong2025mobileworld,cao2025androidlens}, such as AndroidWorld \cite{rawles2024androidworld}, LlamaTouch \cite{zhang2024llamatouch}, and Mobile-Env \cite{zhang2023mobile}. 
Unlike static benchmarks, these environments evaluate task completion based on device state detection or LLM judgments rather than requiring user actions to precisely match predefined answers.
However, these dynamic benchmarks mainly rely on offline static apps (e.g., Google Suite apps, F-Droid apps, and built-in system apps) \cite{liu2025llm}, which are significantly different from mainstream app designs and cannot represent real-world usage scenarios. 
While SPA-Bench \cite{chen2024spa} has made progress by incorporating popular open-source apps, it remains limited. It cannot support apps and tasks that require logging into an account or have high memory usage.
A key limitation of all of these benchmarks is that they rely on simulators rather than real devices, which both significantly limits the scope of tasks and apps and may not accurately reflect performance in real-world environments.
Additionally, the tasks in existing benchmarks are often too simple or too rigid, neglecting the evaluation of the exploration and generalization capabilities of GUI Agents, which are crucial in dynamic environments.

\section{Experiments Settings}\label{appendix_experiments}

\subsection{Evaluation Metrics}\label{appendix_metric}

We define four key metrics to evaluate agent performance:

$\bullet$ Success Rate (SR): A GUI trajectory is considered a successful completion if it meets the entire task success condition and ends with Complete action. This is the core metric of MobileBench-OL. 

$\bullet$ Sub-condition Success Rate (Sub SR): Sub-SR calculates the ratio of matched sub-conditions to the total number of sub-conditions, regardless of the termination reason. As discussed earlier, each task's condition may consist of multiple sub-conditions involving element matching and action matching. We evaluate whether a GUI trace meets these sub-conditions to provide fine-grained insight into task execution.

$\bullet$ Step Ratio: This metric compares the number of steps taken by the agent to the golden steps operated by human ideally. The agent's max step limit is set to triple the golden steps.  Golden steps excludes redundant steps such as pop-ups or loading waits.

$\bullet$ Failure Reasons: This metric is categorized into three types, see Figure \ref{figure_5_1}: (1) Early Termination: The agent reports task finished without completing the task. (2) Overdue Termination: The agent completes the task but keeps operating until max step. (3) Failure: The agent neither completes the task nor reports finished.

\subsection{Implementation Details. }
During the evaluation process, all test phones have all apps pre-installed and account logins completed. After each round of MobileBench-OL evaluation, we will perform  reset tasks to restore the phone to its initial state.

The specific evaluation process is as follows: The GUI agent will cyclically obtain environmental information from the device platform, extract prompts, generate actions, and finally send them back to the platform for execution. This loop will continue until the task is completed or the max step limit is reached. The max step limit for the task is 3 times the human annotated trajectory’s steps, and there is a 3-second waiting interval after each action is executed. We found that most agents could not correctly start the target app (even if the app has been placed on the main screen of the phone). For this reason, we set all tasks to start directly from the app homepage.

\subsection{Baselines}
We evaluate 8 open-source models and 4 closed-source model under unified benchmark setting. All apps are kept in consistent versions and reset to an initial state before testing to ensure repeatability and fair comparisons.
All open source agents were deployed on an 80GB NVIDIA A100 GPU, while closed source models were accessed through their respective APIs. 
All open-source models are of comparable capacities(7B,8B,9B).

For close-source models:

$\bullet$  GPT-4o: We deliberately avoid auxiliary inputs such as SOM or element lists—unlike many previous works—to ensure a strictly screenshot-only evaluation baselines. 

$\bullet$  M3A: Following the setup in \textit{AndroidWorld} \cite{rawles2024androidworld}, M3A uses GPT-4o as the backbone model, taking UI element information and SoM-enhanced screenshots as input to generate actions. It further includes a reflection step—comparing pre- and post-action UI states—and incorporates the reflection summary into the action history for subsequent decisions.

$\bullet$ T3A: T3A adopts a similar pipeline and also uses GPT-4o as the backbone model, but operates in text-only mode: it relies solely on UI element textual information, without any visual input (e.g., screenshots or SoM).

$\bullet$ Mobile-Agent-V2 \cite{wang2024mobile}: Mobile Agent V2 employs a multi-agent pipeline comprising a Planning Agent, a Decision Agent, and a Reflection Agent, all built upon GPT-4o as the backbone model. To support icon understanding, it integrates an external icon captioning model via the Qwen-VL-Plus API. We adapt it to our architecture.

For open-source models:

$\bullet$ Qwen2.5-VL and UI-TARS are fine-tuned from Qwen2-VL; the remaining baselines (e.g., InternVL2-8B, CogAgent-9B) are run at a comparable capacity, primarily 7B with occasional 8B or 9B variants, to ensure fair comparison. 

$\bullet$ We standardize prompting as follows: all Qwen-based variants use the official UI-TARS prompt; models with vendor-provided prompts/templates (e.g., CogAgent, OS-Atlas) use their official settings; all remaining baselines use the AndroidWorld generic prompt.

\section{More Experiments}\label{appendix_more_experiments}

\begin{table*}[!t]
    \centering
\resizebox{1\textwidth}{!}{
    \begin{tabular}{l|cc|cc|cc|cc|cc}
    \hline\hline
        \multirow{2}{*}{Model}   &    \multicolumn{2}{c|}{Base}   &    \multicolumn{2}{c}{Long-Tail}  & \multicolumn{2}{c}{Long-Horizon} & \multicolumn{2}{c}{GUI-Reasoning}  & \multicolumn{2}{c}{Noise-Robust}  \\ \cline{2-11} 
        &   All Tasks & SR Tasks  &   All Tasks & SR Tasks  &   All Tasks & SR Tasks  &   All Tasks & SR Tasks  &   All Tasks & SR Tasks  \\ \hline 
        GPT-4o      &   2.16& 1.55& 2.22& 1.46& 0.88 & - & 1.97 & 1.66 & 2.40 & 1.75 \\ 
        M3A & 1.95	&1.23	& 2.10& 1.39 & 1.15	&0.98	& 2.06& \textbf{1.05} & 2.74	&1.65  \\ 
         T3A & 1.53	&1.56	& 1.40& 1.37& \textbf{0.54}	& \textbf{0.29}	& \textbf{1.40} & 1.43 & 2.41	&1.60  \\ 
         Mobile-Agent-V2 & 2.28	&1.48	& 2.33& 1.63 & 1.11	& 0.47	& 1.97& 1.50 & 2.53	&1.71 \\ \hline
        InternVL2-8B &   1.83 & 2.01 & 2.04 & 2.13 & 1.41 & - & 1.89 & - & 3.23 & -\\
        CogAgent-9B   &   \textbf{1.43} & \textbf{1.38} & 1.24 & \textbf{1.41} & 1.12 & - & 1.48 & - & \textbf{1.76} & 1.59\\
        UGround-V1-7B  &   2.54 & 2.42 & 2.01 & 2.07 & 1.41 & - & 2.49 & - & 3.11 & -\\
        OS-Atlas-Pro-7B    &   2.62& 1.81& \textbf{1.00} & 2.07 & 1.35 & - & 2.45 & 2.17 & 3.22 & 2.00 \\
        Qwen2-VL-7B   &   1.98& 1.51& 1.08 & 1.76 & 0.93 & - & 2.21 & - & 2.77 & \textbf{1.49}\\
       Qwen2.5-VL-7B   &   1.98& 1.49& 2.25 & 1.47 & 0.82 & - & 1.95 & 1.39 & 2.00 & 1.57\\
        UI-TARS-7B   &   1.81 & 1.37 & 2.24 & 1.54 & 1.51 & 1.55 & 1.87 & 1.41 & 2.19 & 1.55\\
        UI-TARS-1.5-7B &   1.73 & 1.41 & 2.13 & 1.51 & 2.11 & 1.31 & 1.86 & 1.43 & 1.97 & 1.51\\\hline\hline
    \end{tabular} }
    \caption{The Step Ratio ($\downarrow$)  performance of GUI agents across all tasks and SR tasks on MobileBench-OL. A dash ("-") indicates that the agent has no successful tasks. "All Tasks" represents the overall step ratio across all tasks, while the "SR Tasks" represents the step ratio for successfully completed tasks. } 
    \label{result_step_1}
\end{table*}

\subsection{Long-Horizon Error Analysis}\label{appendix_long_horizon_error}
We classify the errors of UI-TARS-1.5 on the Long-Horizon subset into five categories.  For each category, we provide a detailed definition and examples below. If a task contains multiple errors, we mark it according to the first error that occurs.

$\bullet$ Reasoning \& Planning Failure:  The agent has a biased understanding of semantics or has formulated an incorrect execution plan. This includes inconsistencies between the trajectory intent and the task intent, or an incorrect execution order of subtasks.

\begin{tcolorbox}[   
title=Reasoning \& Planning Failure,
    colback=cyan!3!white,     
    colframe=cyan!40!black,
breakable, 
                  ]
Error Case 1:

\textbf{Task:} I want to use Amap to view nearby  coffee shops, filter them to those listed for more than 3 years, view the details of the most popular coffee shops, plan public transportation routes to those coffee shops, add them to my favorites, and then go to my favorites to view the routes.

\textbf{Result:}  The agent did not add the selected route to my favorites or go to my favorites; instead, they clicked “Start Navigation” directly to view the real-time navigation status.

\tcblower

Error Case 2:

\textbf{Task:} Send "Good morning" to DingTalk assistant on the DingTalk, and then search for emojis for "french fries," "hamburger," "egg tart," "cookie," and "chocolate," select one of each, and send them to the agent.

\textbf{Result:}  The agent does not search for emojis; instead, it directly sends "french fries," "hamburger," etc., as plain text.

\end{tcolorbox}

$\bullet$ Subtask Omission: The agent completely forgets to complete a subtask within the complex task.

\begin{tcolorbox}[   
title=Reasoning \& Planning Failure,
    colback=cyan!3!white,     
    colframe=cyan!40!black,
breakable, 
                  ]
Error Case 1:

\textbf{Task:} Go to NetEase Cloud Music and add "Selling Newspapers," "Planting the Sun," and "The Barber" to my favorite music list. Then go to my favorite music list to view them.

\textbf{Result:}  The agent did not complete the subtask of searching for "The Barber" and adding it to my favorite music list.

\end{tcolorbox}

$\bullet$ Function Navigation Failure: The agent knows which function it needs, but cannot navigate to the entry point to activate that function within the app.

\begin{tcolorbox}[   
title=Reasoning \& Planning Failure,
    colback=cyan!3!white,     
    colframe=cyan!40!black,
breakable, 
                  ]
Error Case 1:

\textbf{Task:} Go to Flush to check the A-share stock rankings. Browse the gainers list, losers list, and the stock with the highest trading volume respectively, and view their financial disclosures in the "Important Event Reminders" section of their announcements and information sequentially.

\textbf{Result:}  The agent could not find the functional entry points for "Important Event Reminders" and financial disclosures.

\end{tcolorbox}

$\bullet$ Attribute Omission/Error: The agent attempts to complete a subtask, but the trajectory is incomplete or incorrect. Specifically, when completing a subtask that requires multiple forms or attributes, the agent misses one or two attributes or does not fully follow the task instructions. 

\begin{tcolorbox}[   
title=Reasoning \& Planning Failure,
    colback=cyan!3!white,     
    colframe=cyan!40!black,
    breakable, 
                  ]
Error Case 1:

\textbf{Task:} Open Bilibili and search for "Minecraft". Sort the search results by view count and filter for videos published in the past week. Select a video, like it, and add it to my favorites. On its comment page, sort the comments by time, like the top comment, then edit a comment with "haha" and a smiling emoji and post it.

\textbf{Result:}  The agent did not favorite the video, did not sort the comments on the comment page by time, and could not find a smiling emoji.

\end{tcolorbox}

$\bullet$ Visual Grounding Failure: The agent fails to locate or interact with the correct UI elements, including errors such as being unable to locate small elements and incorrectly using sliding components.

\begin{tcolorbox}[   
title=Reasoning \& Planning Failure,
    colback=cyan!3!white,     
    colframe=cyan!40!black,
breakable, 
                  ]
Error Case 1:

\textbf{Task:} Open Bilibili and search for "gameplay live streams." Select a live stream from the live channel to enter. In the player settings of that live stream room, set it to stop playback after 20 minutes. In the bullet chat settings, set the display area to 3/4 of the screen and the bullet chat speed to fast.

\textbf{Result:}  When setting the time to 20 minutes, the agent did not correctly locate the starting point for the swipe action and was unable to successfully set the time element to 20.

\tcblower

Error Case 2:

\textbf{Task:} In Baidu Browser, search for pictures of Huskies, Samoyeds, and Alaskan dogs respectively, download them, and go to "My Pictures" to view them.

\textbf{Result:}  When the agent went to "My Pictures," the app popped up a cloud storage recommendation ad. The agent's ability to locate the small component was insufficient, and it was unable to successfully click the close button, continuing to fail the operation until the trajectory steps were exhausted.

\end{tcolorbox}

\begin{table}[!t]
    \centering
\resizebox{0.8\columnwidth}{!}{
    \begin{tabular}{l|ccc}
    \hline\hline
    \multirow{2}{*}{Model} & \multicolumn{3}{c}{\textbf{pass@k}} \\ \cline{2-4}
     &  k=1 & k=3 & k=5  \\\hline
    Qwen2.5-VL   & 17.10\% & 31.29\% & 39.35\% \\
    UI-TARS      & 46.77\% & 56.45\% & 69.35\% \\
    UI-TARS-1.5  & 60.97\% & 69.03\% & 73.87\% \\\hline\hline
    \end{tabular}}
    \caption{Pass@k  ($k\in\{1,3,5\}$).}
    \label{tab:top12_passk}
\end{table}

\subsection{Step Ratio}\label{appendix_step_ratio}
In Table \ref{result_step_1}, a lower step-ratio indicates higher efficiency. For agents with SR below 5\%, their Step Ratio is better because they only handle simple tasks and terminate complex tasks prematurely. Better-performing agents have better Step Ratio. This is because they are less prone to loops or aimless exploration and thus finish with step counts closer to the golden path.

\subsection{Pass@K Experiment}\label{appendix_passk}
As shown in Table \ref{tab:top12_passk}, pass@k evaluation (at least one success out of k attempts) on the Base subset reveals that UI-TARS-1.5 achieves 69.03\% (a 1.21x gain) with 3 retries and 73.87\% (a 1.29x gain) with 5 retries.
Overall, while all models benefit from retrying, the gains diminish as the number of retries increases.

\end{document}